\documentclass[sn-apa, iicol]{sn-jnl}

\jyear{2021}%

\theoremstyle{thmstyleone}%

\usepackage{tabularx}
\usepackage{physics}
\usepackage{bbm}
\usepackage{amsmath}
\usepackage{array}
\usepackage{graphicx}

\theoremstyle{thmstyletwo}%
\DeclareMathOperator*{\argmax}{arg\,max} 
\theoremstyle{thmstylethree}%
\newcommand{\etal}{{et al}.\@ }
\begin{document}

\title[Article Title]{Improving Domain Adaptation Through Class Aware Frequency Transformation}


\author[1]{\fnm{Vikash} \sur{Kumar}}\email{vikashks@iisc.ac.in}
\equalcont{These authors contributed equally to this work.}

\author[1]{\fnm{Himanshu} \sur{Patil}}\email{hipatil1998@gmail.com}
\equalcont{These authors contributed equally to this work.}

\author[1]{\fnm{Rohit} \sur{Lal}}\email{rohitlal@iisc.ac.in}

\author*[1]{\fnm{Anirban} \sur{Chakraborty}}\email{anirban@iisc.ac.in}
\affil[1]{\orgdiv{Dept. of Computational and Data Science}, \orgname{Indian Institute of Science}}


\abstract{In this work, we explore the usage of the Frequency Transformation for reducing the domain shift between the source and target domain (e.g., synthetic image and real image respectively) towards solving the Domain Adaptation task. Most of the Unsupervised Domain Adaptation (UDA) algorithms focus on reducing the global domain shift between labelled source and unlabelled target domains by matching the marginal distributions under a small domain gap assumption. UDA performance degrades for the cases where the domain gap between source and target distribution is large. In order to bring the source and the target domains closer, we propose a novel approach based on traditional image processing technique \textbf{C}lass \textbf{A}ware \textbf{F}requency \textbf{T}ransformation (\textbf{CAFT}) that utilizes pseudo label based class consistent low-frequency swapping for improving the overall performance of the existing UDA algorithms. The proposed approach, when compared with the state-of-the-art deep learning based methods, is computationally more efficient and can easily be plugged into any existing UDA algorithm to improve its performance. Additionally, we introduce a novel approach based on absolute difference of top-2 class prediction probabilities (ADT2P) for filtering target pseudo labels into clean and noisy sets. Samples with clean pseudo labels can be used to improve the performance of unsupervised learning algorithms. We name the overall framework as \textbf{CAFT++}.
We evaluate the same on the top of different UDA algorithms across many public domain adaptation datasets.  Our extensive experiments indicate that CAFT++ is able to achieve significant performance gains across all the popular benchmarks.}

\maketitle

\section{Introduction}\label{sec1}

Deep learning has achieved significant success in the computer vision tasks and vastly revolutionised the computer vision research. 
However, most of the state-of-the-art approaches need a large amount of labelled data, collection of which is expensive and time-consuming. Deep models trained on one task can be transferred to another related task without requiring large labelled training datasets for the latter task and this area of research is called Deep Transfer Learning  \cite{tan2018survey}. Although Deep learning-based solutions currently provide state-of-the-art results in most computer vision (CV) applications, there are still many open challenges, such as slow convergence, higher inference time, etc. Even in this era of deep learning, traditional computer vision techniques often remain very relevant - either to provide a better alternative or towards enabling the learning-based solutions to mitigate some of the aforementioned challenges. Data augmentation \cite{perez2017effectiveness} is one of the most popular traditional CV approaches, which directly affects the deep learning model performance by improving its generalization ability. Recently, the fusion of traditional CV along with deep models \cite{o2019deep} is getting attention due to the necessary speedups they provide, which is really essential for real-world applications.

Domain Adaptation is a class of ML techniques that aims to bridge the domain gap across datasets when the underlying distributions of source domain (analogous to training data) and target domain (analogous to test data) samples are different i.e $\mathcal{D_S}(X_s,Y_s) \neq \mathcal{D_T}(X_t,Y_t)$ where $\mathcal{D_S}$ is source data distribution and $\mathcal{D_T}$ is target data distribution.

Most of the unsupervised domain adaptation work \cite{ganin2015unsupervised,sun2016deep} try to close the global domain gap, which is the gap between marginal source and target data distributions, without considering the domain gap in label space.  
However, negative transfer \cite{wang2019characterizing} will occur as a result of only the marginal source and target distributions alignment.
\cite{zhao2019learning} also showed that marginal alignment of the source and target distributions does not ensure joint alignment of $\mathcal{D_S}(X_s, Y_s)$ and $\mathcal{D_T}(X_t,Y_t)$ and therefore, the alignment of marginal distributions of the source and the target domain leads to sub-optimal results. Sub-domain adaptation \cite{zhu2020deep} handles this problem by performing both global and local alignments where global alignment implies alignment between marginal (feature space) distribution $\mathcal{D_S}(X_s)$ and $\mathcal{D_T}(X_t)$ alignment and local alignment implies conditional distribution ($\mathcal{D_S}(Y_s/X_s)$ and $\mathcal{D_T}(\hat{Y}_t/X_t)$) alignment, effectively approximating the joint alignment of source and target data distribution $\mathcal{D_S}(X_s, Y_s)$ and $\mathcal{D_T}(X_t, \hat{Y}_t)$ where $\hat{Y}$ represents pseudo labels. Domain alignment becomes more challenging when the domain gap between the source and the target domains is wider. To avoid negative transfer (transfer between the non-similar classes of two domains), our solution uses a class-aware frequency transform to bridge the domain gap between the source and target domains. Our proposed approach relies on target pseudo labels for performing the class-aware transformation. Quantifying uncertainty associated with pseudo labels is really important to accept or reject a pseudo label. Saito \etal \cite{saito2017asymmetric} quantifies sample certainty using consistency between two neural network outputs. Since it requires additional neural network, it can not be easily integrated to the existing UDA approaches. Wang \etal \cite{wang2021uncertainty} proposes entropy to estimate uncertainty in neural network prediction for semantic segmentation task. Uncertainty quantification through entropy may be affected by a higher number of classes which is often the case for the classification task compared to segmentation, where we usually have fewer classes. Therefore, estimating the uncertainty through entropy can be helpful for segmentation but may not reflect the same for classification task. Therefore, we propose to use the absolute difference between top-2 prediction probabilities (ADT2P) followed by fitting a two component Gaussian Mixture Model (GMM) for quantifying the uncertainty associated with pseudo labels, which we use to filter clean and noisy pseudo labels. We draw the motivation for the same from Figure \ref{fig:pseudo_label} in which we can see that the 2-component GMM can separate clean and noisy pseudo labels when applied over ADT2P. We can use the clean pseudo labels to fine-tune and improve target adaptation performance. We compare our proposed approach by replacing ADT2P with entropy and observe that ADT2P performs better than entropy for all the tasks associated to Office-Home dataset (Table. \ref{tab:top2_vs_entropy}).

\begin{figure}[ht]
\includegraphics[scale=0.38]{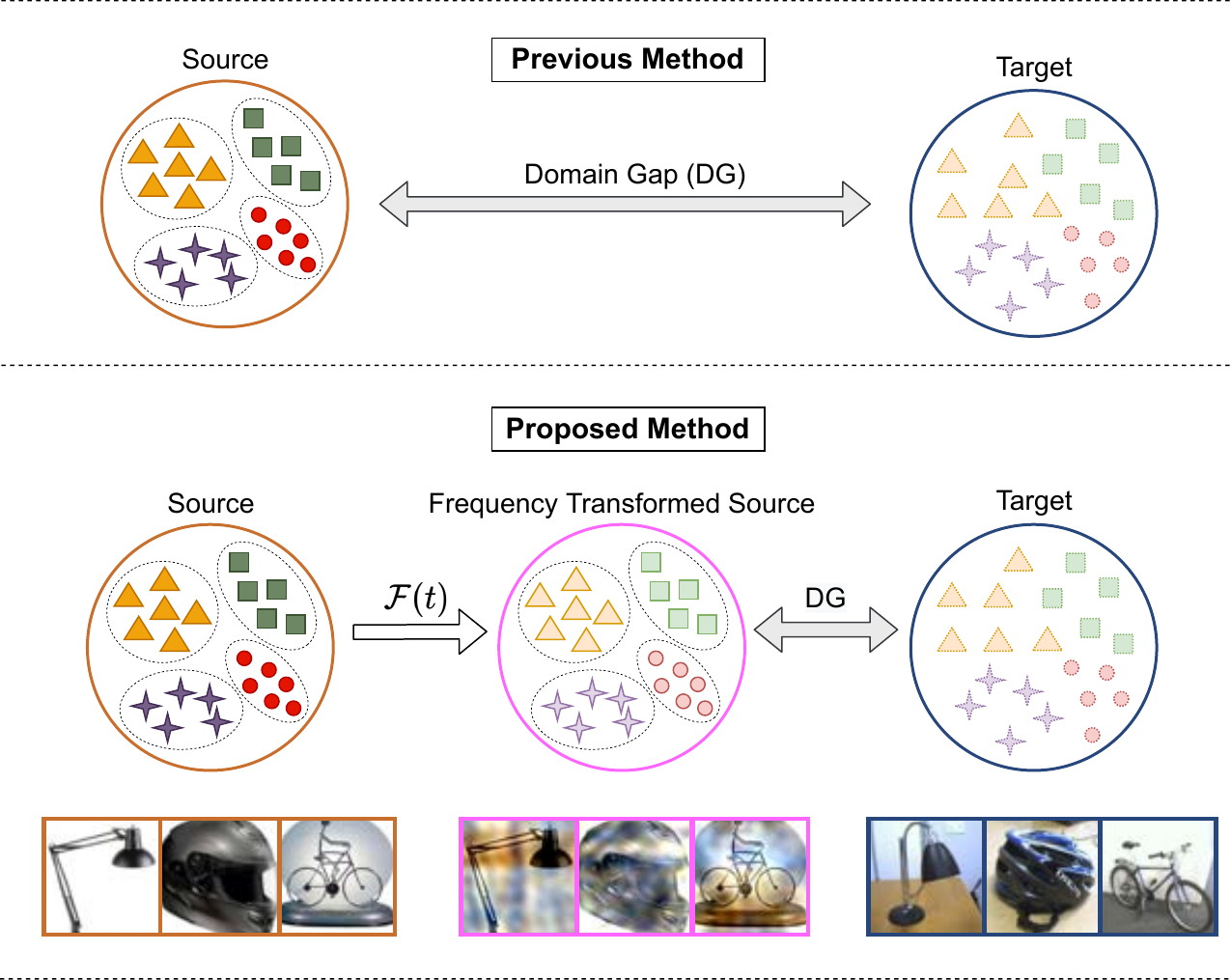}
\label{fig:overview_img}
\caption{\small \textbf{Method Overview}. (\textit{Top Row}) represents general Unsupervised Domain Adaptation (UDA) setting with labelled source images and unlabelled target images. UDA algorithms try to reduce the domain gap. (\textit{Bottom Row})  Our proposed method explicitly tries to swap the source image style with that of the target image using class aware frequency transform towards improved transferability. }
\end{figure}

Transferring the target low-frequency component (style) to source during domain adaptation via Fourier transformation overcomes the domain gap and increases adaptation for segmentation tasks, according to Fourier Domain Adaptation (FDA) \cite{yang2020fda}. Directly extending the FDA for classification problems raises two challenges. Firstly, since the samples from the target datasets are chosen at random, it is vulnerable to negative transfer because it overlooks the notion that within-class domain shift differs from global shift between two domains. Secondly, the artefacts such as image blur are common in modified source dataset generated via inverse Fourier transformation (target style is applied over both background and foreground of the source image), resulting in weaker class-discriminative features. We propose pseudo label based class-aware alteration of the source samples to mitigate negative transfer and reduce the sub-domain gap, which addresses the first problem. To address the second issue, we propose that both original source samples and transformed source samples be used for training because it can compensate for the loss of class-discriminatory characteristics appearing due to low-frequency swapping on the source dataset. As a result, the proposed approach lowers the domain shift explicitly. The proposed technique is depicted in Figure \ref{fig:overview_img}. It assists adaptation strategies in improving their performance because of the improved transferability.  

In summary, our proposed framework comprises the following key contributions - 

\begin{itemize}
\item A novel computationally inexpensive approach that limits the negative transfer and explicitly tries to swap the source image style with that of the target image using class aware Fourier transform for improved transferability.
\item A novel pseudo label filtering strategy based on absolute difference between top-2 prediction probabilities (ADT2P) and Gaussian mixture model (GMM), which separates sample prediction probabilities into clean and noisy pseudo labels. Clean pseudo labels help to perform class-aware sample selection towards transferring style from the target domain to the source.
\item We propose to utilize both the original and the transformed samples to account for the class-discriminative features lost during style transfer.
\item We perform extensive experiments and ablation to showcase the efficacy of our proposed approach. We also show that the proposed solution works well for UDA with category or label shift along with the standard close set UDA.
\end{itemize}

\section{Related Work}
\label{sec:related_work}
\textbf{Unsupervised Domain Adaptation (UDA): } UDA approaches \cite{wilson2020survey} attempt to learn domain invariant representations using a variety of strategies, including reducing divergence, adaptation through reconstruction, and adversarial training methods. Methods like Maximum mean discrepancy (MMD) \cite{gretton2006kernel}, correlation alignment (CORAL) \cite{sun2016return}, contrastive domain discrepancy \cite{kang2019contrastive}, and other techniques try reducing the distance between the statistics of the source domain distribution and the statistics of the target domain distribution. The aim of these strategies is to reduce the domain gap. As these approaches attempt to align the marginal distributions, they are prone to yielding sub-optimal results. DRCN \cite{ghifary2016deep} tries adaptation through reconstruction, which optimizes for a latent  representation that can classify the source domain and reconstruct the target jointly. It will ensure that both the source and target domains have meaningful latent representations. Recent works by \cite{lalopen}, \cite{roy2021curriculum} leverage the use of Graph Convolution Networks to mitigate the domain gap between source and target domain. 

\noindent\textbf{Label Shift in UDA: } In vanilla UDA, we assume that the number of classes are same in both source and target domains (Close-set UDA) therefore, label or category shift does not exist. Label shift happens in UDA when there are private classes in either source domain (Partial-set UDA) or both source and target domain (Open-set UDA), along with common classes between these two domains. This makes the standard UDA more challenging because, while performing target adaptation, the algorithm should be able to handle the label shift as well. Cao \etal \cite{cao2018partial} propose to use multiple domain discriminators with class and instance-level weighting mechanisms to achieve per-class adversarial distribution matching to solve partial domain adaptation. Cao \etal \cite{cao2018partial2} improve on partial domain adaptation by using only one adversarial network and integrating class-level weighting on the source classifier. Zhang \etal \cite{zhang2018importance} proposes a probabilistic weighting scheme using additional adversarial network to find source samples that are similar to the target samples. Busto \etal \cite{panareda2017open} introduced open set domain adaptation where classes private to source and target domain are considered unknown. They map target samples to source classes using an Assign-and-Transform-Iteratively (ATI) algorithm and then train SVMs for final classification. We validate the efficacy of our proposed approach in label shift condition and observe that it improves the existing performance.

\begin{figure*}[!t]
\centering
\includegraphics[scale=.465]{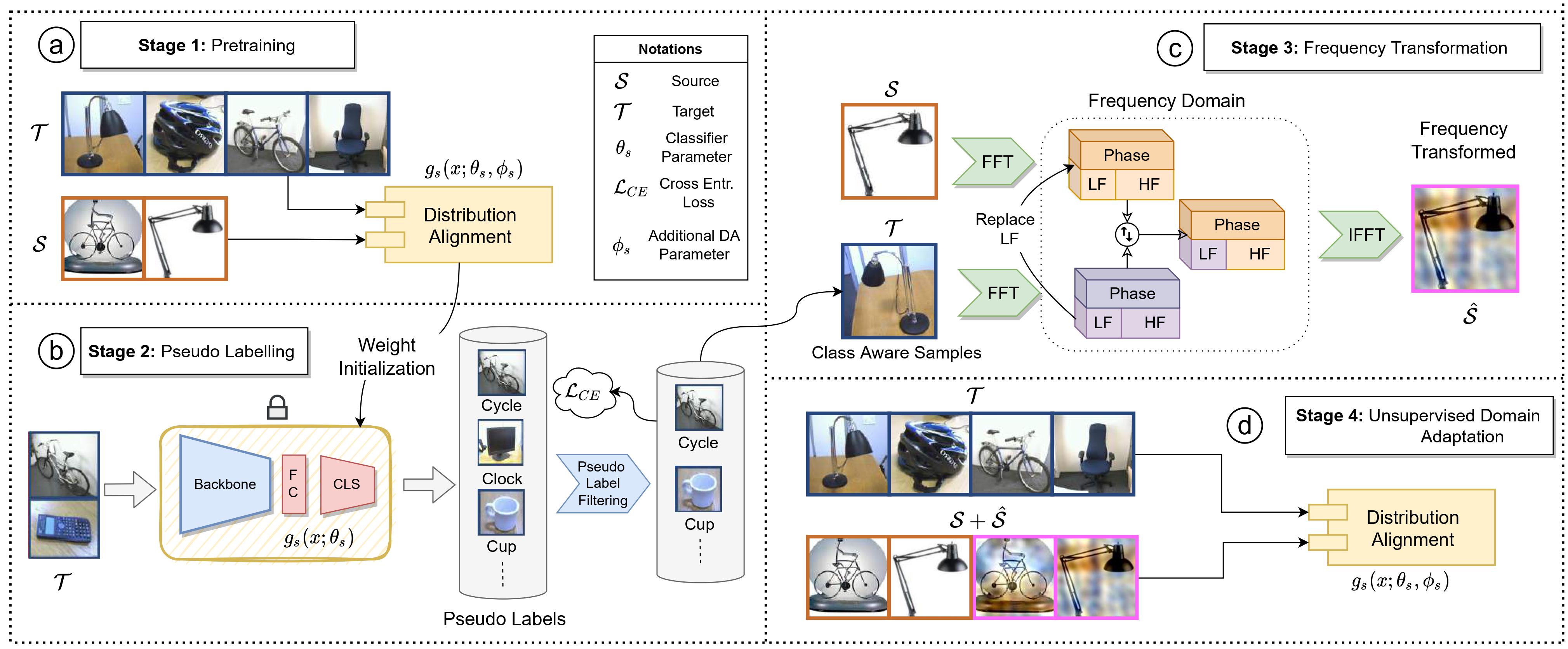}
\label{fig:architecture}
\caption{\small \textbf{Architecture of CAFT++}. The architecture is divided into four stages. In Stage 1, we train a domain adaptation network using the labelled source domain dataset and unlabelled target domain dataset. This trained model is then used to generate pseudo labels in Stage 2. We pass the image through the network to get the difference in top-2 prediction probabilities. We further process this difference in probabilities through a GMM to decide whether it is clean or noisy. In Stage 3, we transform the source image using frequency domain manipulation with the help of the generated target pseudo labels. The transformed source is closer to the target. The source $\mathcal{S}$, Transformed source $\mathcal{\hat{S}}$ (with labels), and target $\mathcal{T}$ images (unlabelled) are then passed through the Domain Adaptation network in Stage 4.}
\end{figure*}

\noindent\textbf{Image Data Augmentation:}  Data augmentation is a widely adopted method when it comes to reducing the domain shift that exists between the source and target domain. Methods such as  norm-VAE \cite{wang2020data} and AdaIN \cite{huang2017arbitrary} try to use learning-based style transfer in order to implement augmentation based domain gap reduction. These techniques pose a cost in terms of computational power and the training time. Existing GAN based methods like CycleGAN \cite{zhu2017unpaired} requires substantial computational cost and takes a lot of time to converge. On the other hand, our proposed method CAFT doesn't depend on any trainable parameter and is relatively free of hyperparameters. Hence, it is fast and computationally inexpensive compared to generative methods.

\noindent\textbf{Pseudo Labeling:} Pseudo labelling was initially used in semi-supervised learning as an entropy regularisation approach, in which unlabeled samples were allocated to the class with the highest output probability \cite{lee2013pseudo}. Because of its effectiveness, the pseudo label based refinement has been embraced in many unsupervised tasks, including unsupervised domain adaptation. From classic marginal distribution alignment \cite{ganin2015unsupervised} to conditional distribution alignment \cite{long2017deep,long2017conditional}, pseudo label assisted in improving domain adaptation algorithms. Along with correct ground truth, pseudo labels will also have incorrect labels, which is why they are referred to as noisy pseudo labels. \cite{liang2020we} introduces a $K$-means based pseudo label search to handle the noisy pseudo label. One popular approach to dealing with noisy labels is to utilise loss as a metric for quantifying correct and incorrect labels \cite{li2020dividemix}. Pan \etal \cite{pan2019transferrable} utilises nearest class prototypes for assigning pseudo labels to the unlabeled target samples. Later, they calculate class prototype for source-only, target-only and source-target combinations and enforce similar classes to be close in the feature space. Similar to \cite{pan2019transferrable}, Chen \etal \cite{chen2019progressive} uses source prototypes for assigning pseudo label based on the maximum cosine similarity with the assigned class prototype. Additionally, they also use curriculum leaning based training by sorting easy to hard target samples using the similarity score and perform progressive alignment between source and target prototypes. Choi \etal uses clustering density to decide easy and hard samples for curriculum learning and assumes that samples coming from high density clusters are more likely to have correct pseudo labels. 
We propose to use absolute difference of top-2 predicted probabilities as a measure of uncertainty and filter correct and wrong labels under the assumption that difference between top-2 probabilities should be higher for correct sample and lower for incorrect sample. 

\noindent\textbf{Uncertainty Quantification:} Deep neural network predictions play an important role in data-driven decision making, but these predictions often suffer from over and under confidence therefore, quantifying uncertainties is important in those situations \cite{gawlikowski2021survey}. One popular approach for quantifying uncertainty is based on entropy measure where low entropy implies prediction with higher certainty and vice-versa. Wang \etal \cite{wang2021uncertainty}  propose to use entropy for quantifying uncertainty in solving semantic segmentation task. Saito \etal \cite{saito2017asymmetric}  uses consistency between two neural network outputs to quantify certainty associated with the sample. It requires additional parameters and can not be easily integrated into existing methods. Quantifying uncertainty through entropy measure may be challenging when the number of classes are higher. Our proposed approach uses absolute difference between top-2 prediction probabilities and fits a two component Gaussian Mixture Model (GMM) \cite{reynolds2009gaussian} in order to filter clean and noisy samples.

\begin{figure}[!h]
\centering
\includegraphics[width=0.98\linewidth]{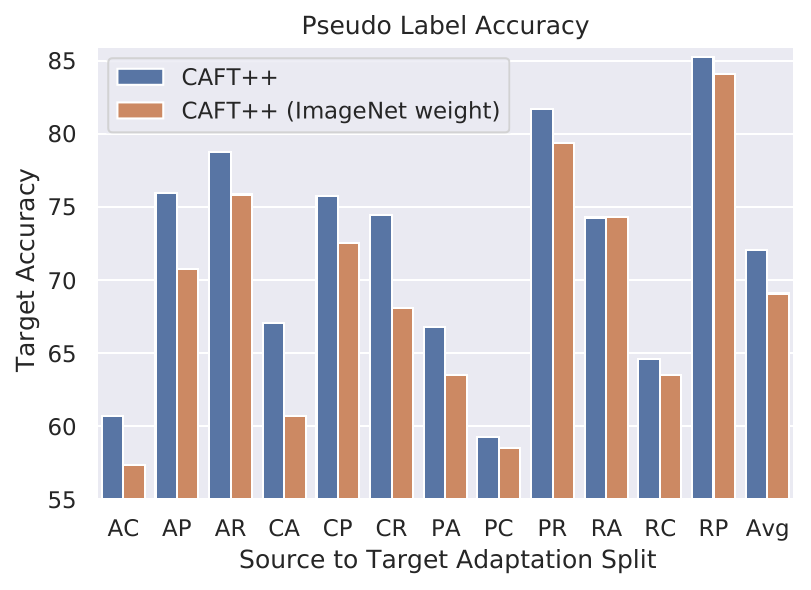}
\caption{\small This ablation shows the effect of weight initialization during warmup stage.}
\label{fig:abla_weight}
\end{figure}

\noindent\textbf{Comparison with CAFT:} We address a major challenge associated with the reliability of pseudo labels in CAFT \cite{Kumar_2021_ICCV} through uncertainty quantification. In CAFT++, we propose a novel method for uncertainty quantification based on the absolute difference between the top-2 prediction probabilities (ADT2P) of unlabelled target samples and the two-component Gaussian mixture model (GMM). It separates pseudo labels into noisy and clean sets. We use these clean pseudo labels for low-frequency magnitude spectrum swapping between the source and the target samples in a class-aware manner. In CAFT, we initiate the training from scratch using ImageNet trained weights. We observe that this strategy leads to the generation of wrong pseudo labels at the start of training. To mitigate this problem, we use pre-trained domain adaptation network weights in CAFT++. These trained weights serve as better initialization compared to pre-trained ImageNet weights. We further validated this hypothesis empirically and showed the results in Figure \ref{fig:abla_weight} of the paper. As expected, we observed performance improvement when we initialized the weights using pre-trained domain net weights. We have analyzed the effectiveness of the proposed class-aware low-frequency transformation of the source sample in CAFT++ by comparing it against MixUp [2], one popular image space augmentation strategy (Section 5.3). We observe that the proposed augmentation strategy in CAFT++ performs better than MixUp (Figure \ref{fig:abla_mixup}). Hence we can say that the augmentation strategy of CAFT++ is better than other similar augmentation strategies, which requires both source and target domain dataset. Additionally, we perform extensive experiments for analysing various components. We validate CAFT++ generalization ability for UDA under category or label shift and show that it works well even under label shift situation.

\begin{figure*}[!]
\centering
\includegraphics[scale=.52]{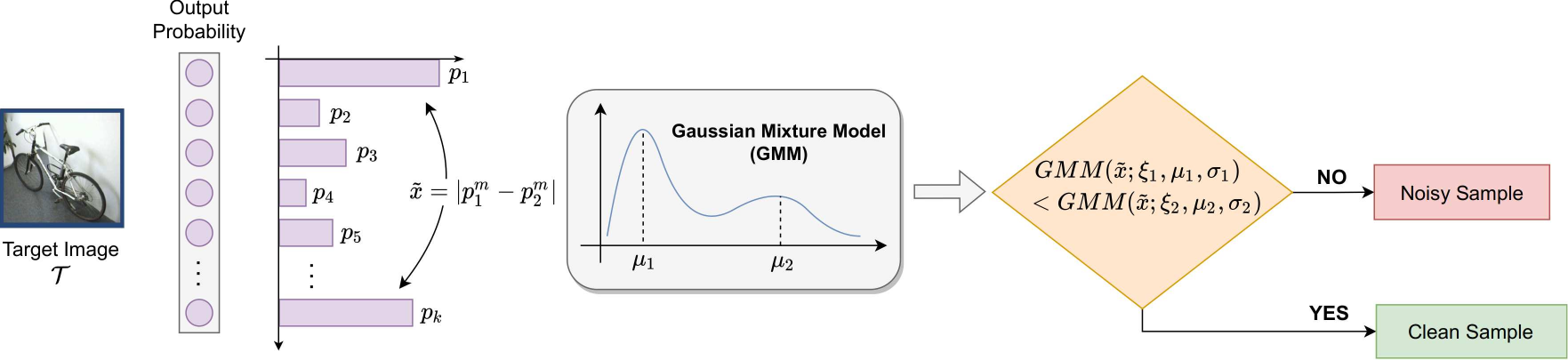}

\caption{\textbf{Method of separating clean and non-noisy pseudo labels}. To reduce the noise content, pseudo labels are filtered using absolute difference of top-2 probabilities (ADT2P). In this method we obtain the predicted probability distribution of an image by passing it through the classifier backbone. We then take the absolute difference $\tilde{x}$ of top-2 probabilities $\tilde{x}=\abs{p_1^m - p_2^m}$ and process it through the Gaussian Mixture Model. If the model predicts that the pseudo label is clean then it gets used for further training, else the sample is discarded. }
\label{fig:ADT2P}
\end{figure*}

\section{Proposed Method}
\label{sec:problem_definition}

We present a four-stage algorithm in Figure \ref{fig:architecture} for deducing domain shift. In Stage-1, we train an existing domain adaptation network and save the trained weights. In Stage-2, we calculate the target prediction probabilities using the trained weights in stage-1. Then we filter out the clean and noisy labels using the absolute difference between top-2 prediction probabilities and Gaussian mixture models. In Stage-3, we do the low-frequency swapping using the class-aware frequency transformation. Finally, in Stage-4, we perform target adaptation and fine-tuning. In the sub-sections below, we will go through each of these steps in further depth.

\subsection{Warm-up with existing Domain Adaptation Network}

Stage-1 is warm-up stage, where we train the existing domain adaptation network as shown in Figure  \ref{fig:architecture}(a). We have access to labelled source dataset and unlabelled target dataset for performing target adaptation. In CAFT \cite{Kumar_2021_ICCV}, we observe that the target adaptation performance depends on the quality of pseudo labels. Therefore, we propose to use trained weights using standard domain adaptation network. It helps us to get rid of extremely noisy samples during the start of the training. Though it helps to reduce the noise content in the pseudo label, we still need to filter out noisy and clean samples from the pseudo labels of target dataset for improved performance. In the next stage, we will discuss the novel way to separate noisy and clean samples.

\begin{figure}[!htb]
\centering

\includegraphics[width=8.5cm,height=6.5cm]{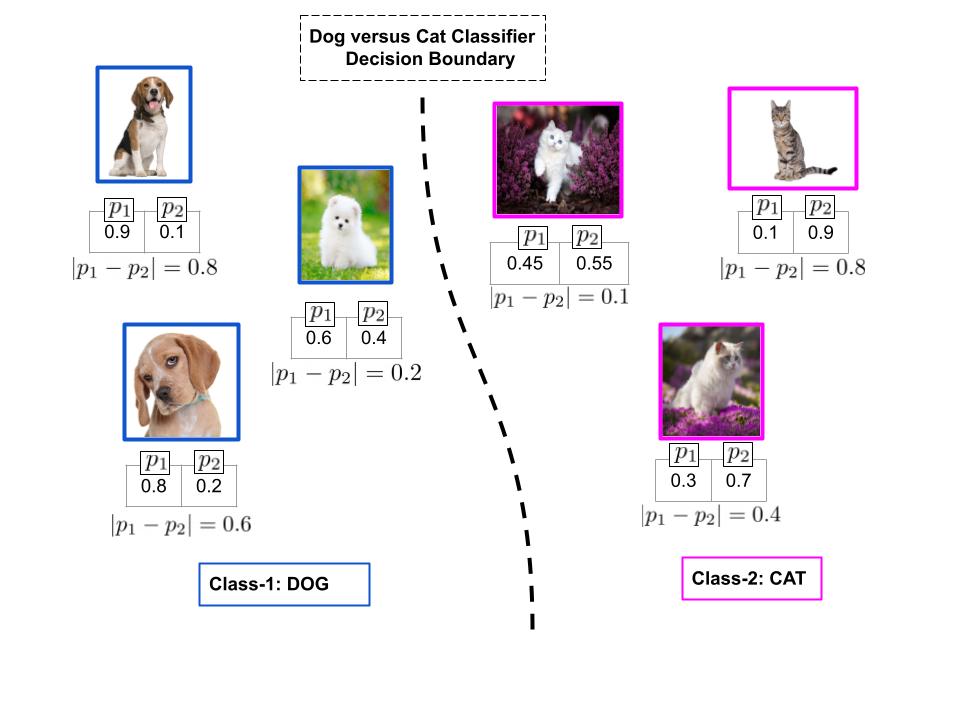}

\caption{\small \textbf{Illustration of absolute difference of top-2 prediction probabilities}. samples close to the decision boundary will have smaller prediction probability compared to samples that are far from the decision boundary and therefore the nearest class probability will also be significant. Hence, the absolute difference between top-2 prediction probabilities will be smaller for a sample near to the decision boundary and higher for the samples which are far from the decision boundary.}

\label{fig:dog_cat}

\end{figure}

\subsection{Separating Clean and Noisy Pseudo Labels}
In this iterative stage, we initiate the target pseudo label calculation using the trained adapted network in Stage-1. We design a novel solution using the absolute difference between top-2 prediction probabilities (ADT2P) and fitting a two component Gaussian Mixture Model (GMM).\\

\begin{figure*}[!]

\centering
\includegraphics[width=0.8\textwidth]{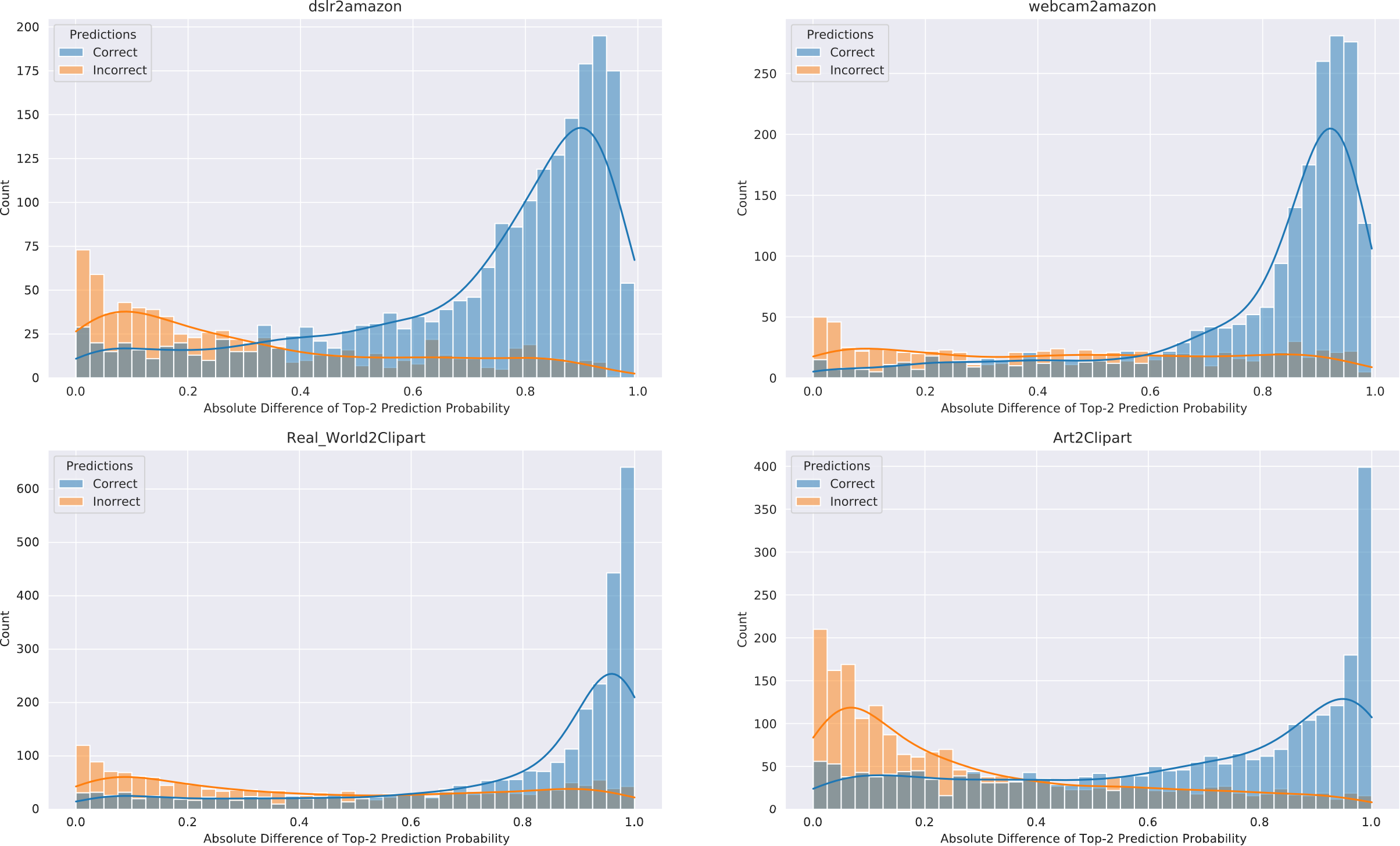}

\caption{\small \textbf{Histogram plot of Absolute difference of top 2 prediction probabilities for various datasets and splits}. We plotted the histogram of absolute difference between top-2 prediction probabilities. We fit the two component Gaussian Mixture Model to separate correct and incorrect pseudo labels. Here, we have used the ground truth only for the representation purpose. We can see that most of the correct pseudo labels are having higher ADT2P value and lie toward the right of the histogram. Similarly, the ADT2P is smaller for the incorrect samples and lies towards the left side. So, if a sample's ADT2P belongs to the Gaussian with higher mean, it will be one of the samples with high confidence and vice-versa. }

\label{fig:pseudo_label}
\end{figure*}

\noindent\textbf{Absolute Difference between Top-2 Prediction Probabilities (ADT2P):} We provide the illustration of overall design in Figure  \ref{fig:dog_cat}. The decision boundary should pass through the low density region for better classification performance. Therefore, when the sample is far from the decision boundary then the entropy will be low (i.e the likelihood of sample belonging to only one of the class is very high, leftmost and rightmost samples in Figure  \ref{fig:dog_cat}) whereas when the sample is nearby the decision boundary then the entropy will be higher (i.e the likelihood of sample belonging to more than one class will be comparable, samples near decision boundary in Figure  \ref{fig:dog_cat}). We hypothesize that the absolute difference between top-2 prediction probabilities are higher for the samples which are far from the decision boundary (\textit{clean samples}) compared to samples that are nearby the decision boundary (\textit{noisy samples}). We draw the motivation for the same from Figure \ref{fig:pseudo_label}. Please refer to Figure  \ref{fig:ADT2P} and algorithm \ref{alg:algorithm} for complete pseudo label filtering procedure. \\

\noindent\textbf{Gaussian Mixture Model (GMM):} Suppose we are given $n$ data-points $\{x^{(1)}, x^{(2)},...,x^{(n)}\}$. We calculate the absolute difference of top-2 prediction probabilities for $n$ data-points and represent as $\{\tilde{x}^{(1)}, \tilde{x}^{(2)}, ..., \tilde{x}^{(n)}\}$. We do not know the ground-truth of these $n$ data-points because target domain is unlabelled in unsupervised domain adaptation. We want to model $p(\tilde{x}^{(i)}, z^{(i)})=p(\tilde{x}^{(i)} \| z^{(i)})p(z^{(i)})$. $z^{i} \sim Multinomial(\xi)$ where $\xi_{j} \geq 0$ and $\sum_{j=0}^{j=\hat{k}}(\xi_{j}) = 1$. Please note that the parameter $\xi_j$ gives the probability that $p(z^{(i)})=j$ and $(\tilde{x}^{(i)} \| z^{(i)}=j)$ is sampled from a normal distribution $\mathcal{N}(\mu_{j}, \Sigma_{j})$, where $\mu_{j}, \Sigma_{j}$ are $j^{th}$ Gaussian distribution parameter. Let $\hat{k}$ is number of values $z^{(i)}$ can take. In our case $\hat{k}$ is two because we will have two components in our Gaussian. Thus, we can say that each $\tilde{x}^{(i)}$ can come from one of the $\hat{k}$ Gaussian (two in our case). $z^{(i)}$ is known as latent random variable which indicate that which of the $k$ Gaussian's each $\tilde{x}^{(i)}$ has come from. We need to estimate the model parameter $\xi, \mu ~\text{and} ~\Sigma$ using the $n$ absolute difference of prediction probabilities. We can write the likelihood equation as shown below:
\begin{align*}
    l(\xi, \mu, \Sigma) &= \sum_{i=1}^{n}\log p(\tilde{x}^{(i)};\xi, \mu, \Sigma) \\
                        &= \sum_{i=1}^{n}\log \sum_{z^{(i)}=0}^{1} p(\tilde{x}^{(i)}\|z^{(i)}; \mu, \Sigma)p(z^{(i)};\xi)
\end{align*}
We can not find a closed form solution of $l(\xi, \mu, \Sigma)$ unless we assume some initial values for $z^{(i)}$. Therefore, expectation and maximization strategy is adopted to find the estimates of model parameters. 

\noindent\textbf{E-Step:} $W_{j}^{(i)} = p(z^{(i)} =j \|x^{(i)};\xi, \mu, \Sigma)~~ \forall i,j$
\noindent\textbf{M-Step:} Update parameters:
\begin{align*}
    \xi_{j} &= \frac{1}{n}\sum_{i=1}^{n} W_{j}^{i} \\
    \mu_{j} &= \frac{\sum_{i=1}^{n}W_{j}^{(i)} \tilde{x}^{(i)}}{\sum_{i=1}^{n}W_{j}^{i}}\\
    \Sigma_{j} &=\frac{\sum_{i=1}^{n}W_{j}^{(i)} (\tilde{x}^{(i)} - \mu_{j})(\tilde{x}^{(i)} - \mu_{j})^{T}}{\sum_{i=1}^{n}W_{j}^{i}}
\end{align*}

\begin{algorithm*}[!htpb]
   \caption{\small{CAFT++: Improving Domain Adaptation through Class Aware Frequency Transformation}}\label{alg:algorithm}
     \textcolor{Maroon}{\underline{\textit{Separating Clean and Noisy
Pseudo Labels}}}
    
    \textbf{Input:} Target domain data $\mathcal{D}_{t}$; Maximum Epoch  $E_{2}^{max}$; Pre-trained model. Dictionary for mapping from pseudo label $pl$ to target sample: $Dict:\{Key:Val\}$; N: Total number of target samples.
    
    \textbf{Initialization:} Network remains frozen; $g_{tj}(x;\theta_{t}) \leftarrow g_{s}(x;\theta_{s})$
    \begin{algorithmic}
        \While{$B < B^{max} $}
        \State {~~$\{x_{i}^{t}\}_{i=1}^{B} \sim \mathcal{D}_{t}$} \Comment{\textcolor{OliveGreen}{\textit{sample batch of target image}}}
        \State{$\{\tilde{x}_{i}^{t}\}_{i=1}^{B}=\{\abs{(p_{i1}^m - p_{i2}^m)}\}_{i=1}^{B}$} \Comment{\textcolor{OliveGreen}{\textit{calculate the absolute difference between top-2 prediction probability}}}
        \EndWhile
    \end{algorithmic}
    Fit the 2 component GMM on $\tilde{x}=\{\tilde{x}_i^t,...,\tilde{x}_N^t\}$, and get the parameters $\mu_1, \Sigma_1, \xi_1, \mu_2, \Sigma_2, \xi_2$
    \begin{algorithmic}
        \While{$i < N $}
            \State {~~$x_{i}^{t} \sim \mathcal{D}_{t}$}
                \If{$GMM(\tilde{x}_{i};\mu_1, \Sigma_1, \xi_1) < GMM(\tilde{x}_{i};\mu_2, \Sigma_2, \xi_2)$} \Comment{\textcolor{OliveGreen}{\textit{Potential Clean sample}}}
                    \State{$y^{pl}_{i} =  \argmax\{g_{t}(x_{i}^{t};\theta_{t})$}
                    \State{Update $Dict:\{Key:Val\}$ with Key $=y_{i}^{pl}$, Val $=$ associated target sample }
                \Else
                    \State{Ignore the sample. It can be noisy} \Comment{\textcolor{OliveGreen}{\textit{Potential Noisy sample}}}
                \EndIf
        \EndWhile
    \end{algorithmic}

    \textcolor{Blue}{\underline{\textit{Frequency Transformation of Source Sample}}}
    
    \textbf{Input:} $\{x_{i}^{s}, y_{i}^{s}\}_{i=1}^{B} \sim \mathcal{D}_{s}$ \Comment{\textcolor{OliveGreen}{\textit{Sample a batch of source image}}}
    
    $\{x_{i}^{t}, y_{i}^{pl}\}_{i=1}^{B} \sim Dict(Key, Val)$ where $y_{i}^{s} = y_{i}^{pl}$\Comment{\textcolor{OliveGreen}{\textit{Sample a source label conditioned batch of target image from dictionary}}}

    Find Fourier transform of source sample: $\mathcal{F(\mathcal{S})} =  \mathcal{F}(x_i^s)$
    
    Find Fourier transform of selected target sample: $\mathcal{F(\mathcal{T})} = \mathcal{F}(x_i^t)$ \Comment{\textcolor{OliveGreen}{Make Sure that $y_{i}^{s}=y_{i}^{pl} ~\forall ~i$ (sample index)}}
    
    Swap source low frequency with that of target: $\tilde{f}\{\mathcal{F(\mathcal{S})}, \mathcal{F(\mathcal{T})}\}$ \Comment{\textcolor{OliveGreen}{\textit{Figure \ref{fig:architecture}(c)}}}

    Get the Transfromed source using inverse Fourier Transform:
    $\hat{\mathcal{S}} = \mathcal{F}^{-1} [ \tilde{f}\{\mathcal{F(\mathcal{S})}, \mathcal{F(\mathcal{T})}\} ]$
    \Comment{Refer Eq. \ref{eq:eq2}}
        
\end{algorithm*}
\subsection{Frequency Transformation of Source Sample}
\textbf{Notation}: We provide the brief details of the notations used in this work. $\mathcal{S}, ~\mathcal{T}, ~\hat{\mathcal{S}}$ are samples from source distribution, samples from target distribution and samples from frequency transformed source distribution respectively. $F[k,l], ~f[m,n], ~\mathcal{F}, ~\mathcal{F}^{-1}$ are element in frequency domain at $[k,l]^{th}$ index, element in image space at $[m,n]^{th}$ index, Fourier transform and inverse Fourier transform respectively.

Class discriminative representations are domain invariant in the sense that these representations are consistent across domains. Domain variant representation, on the other hand, collect domain-related information and vary across the domain. Domain shift occurs mostly as a result of domain variation traits. Because image style is a domain-dependent component, it contributes to domain shift. We deliberately aim to change the source style to the target style in our suggested strategy to decrease the domain shift between transformed source and target data distributions.

\begin{figure*}[t]
    \centering
    \begin{minipage}{0.5\textwidth}
        \centering
        \includegraphics[width=1\linewidth]{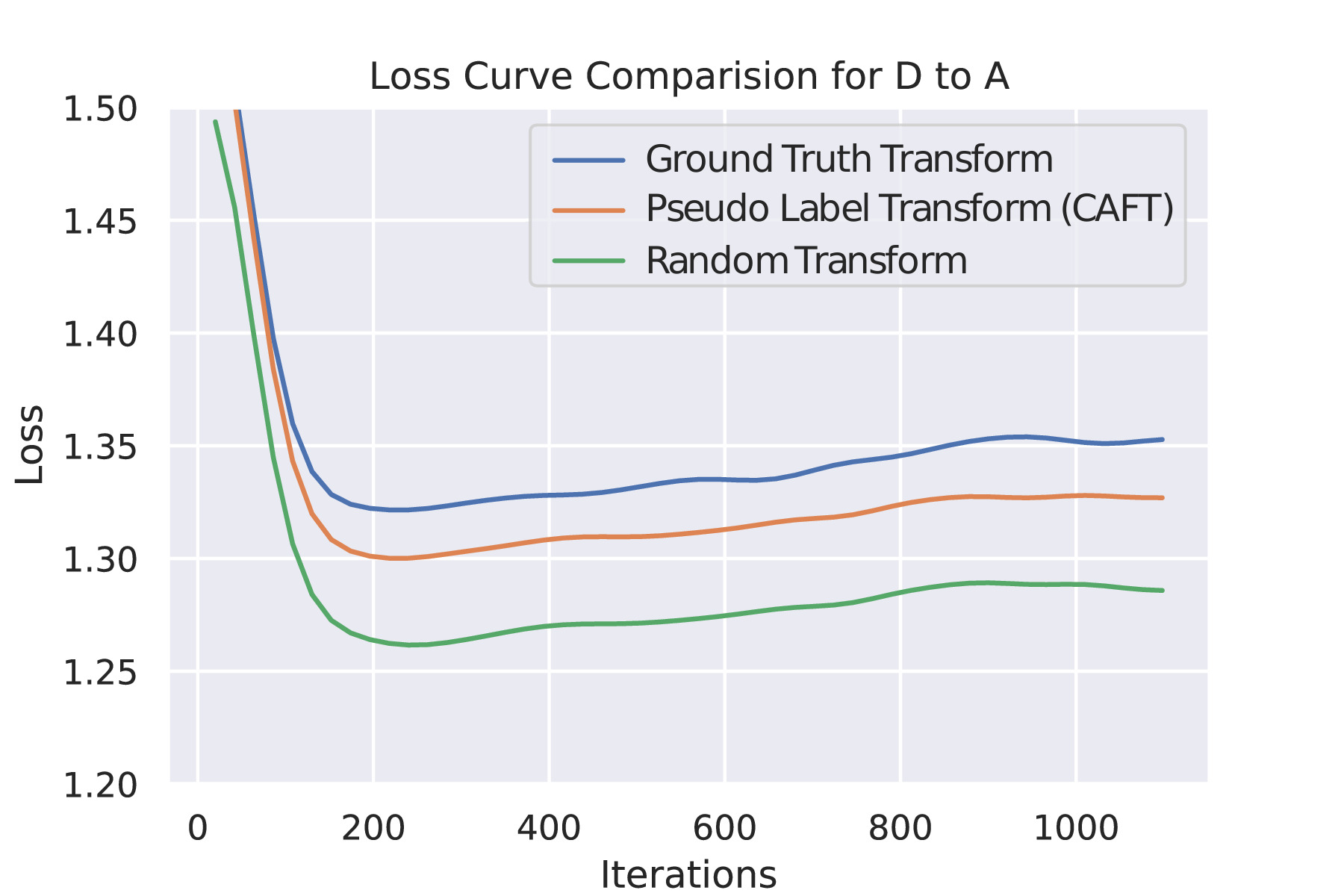}
    \end{minipage}%
    \begin{minipage}{0.5\textwidth}
        \centering
        \includegraphics[width=1\linewidth]{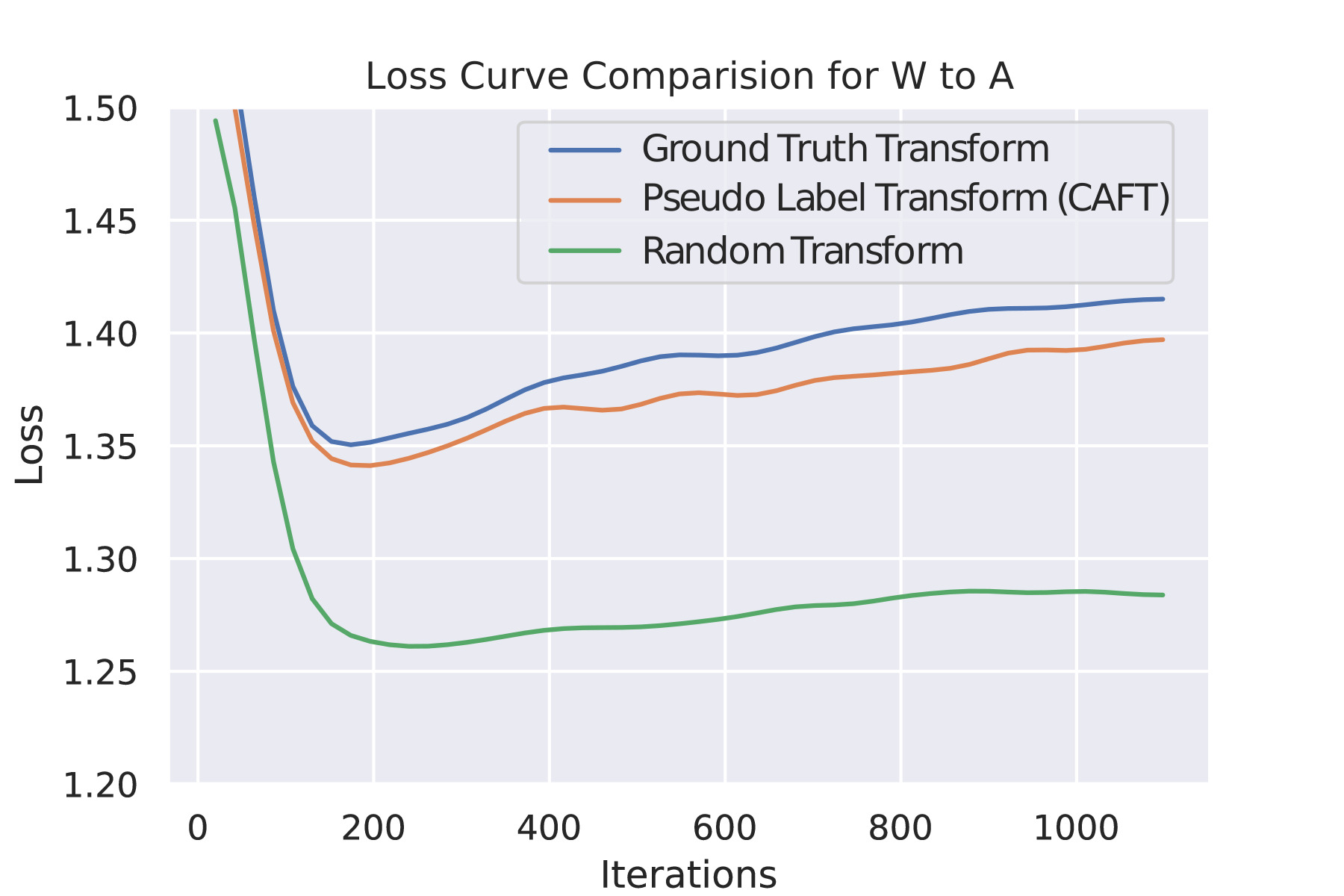}
    \end{minipage}%
    \vspace{1mm}
    \caption{The loss plots for ideal case ground truth based target transformation against the  pseudo label based transformation (CAFT) and random transformation method for DeepCoral \cite{sun2016deep}. The convergence of loss for the CAFT is faster compared to the random transformation based approach. ground truth based target transformation converges fastest, which is in line with our proposed hypothesis. The superior performance of the proposed method against the randomly selected image based transformation method is evident from the graphs.}
    \label{fig:loss_curve}
\end{figure*}

\begin{figure*}[t]
    \centering
    \begin{minipage}{0.5\textwidth}
        \centering
        \includegraphics[width=1\linewidth]{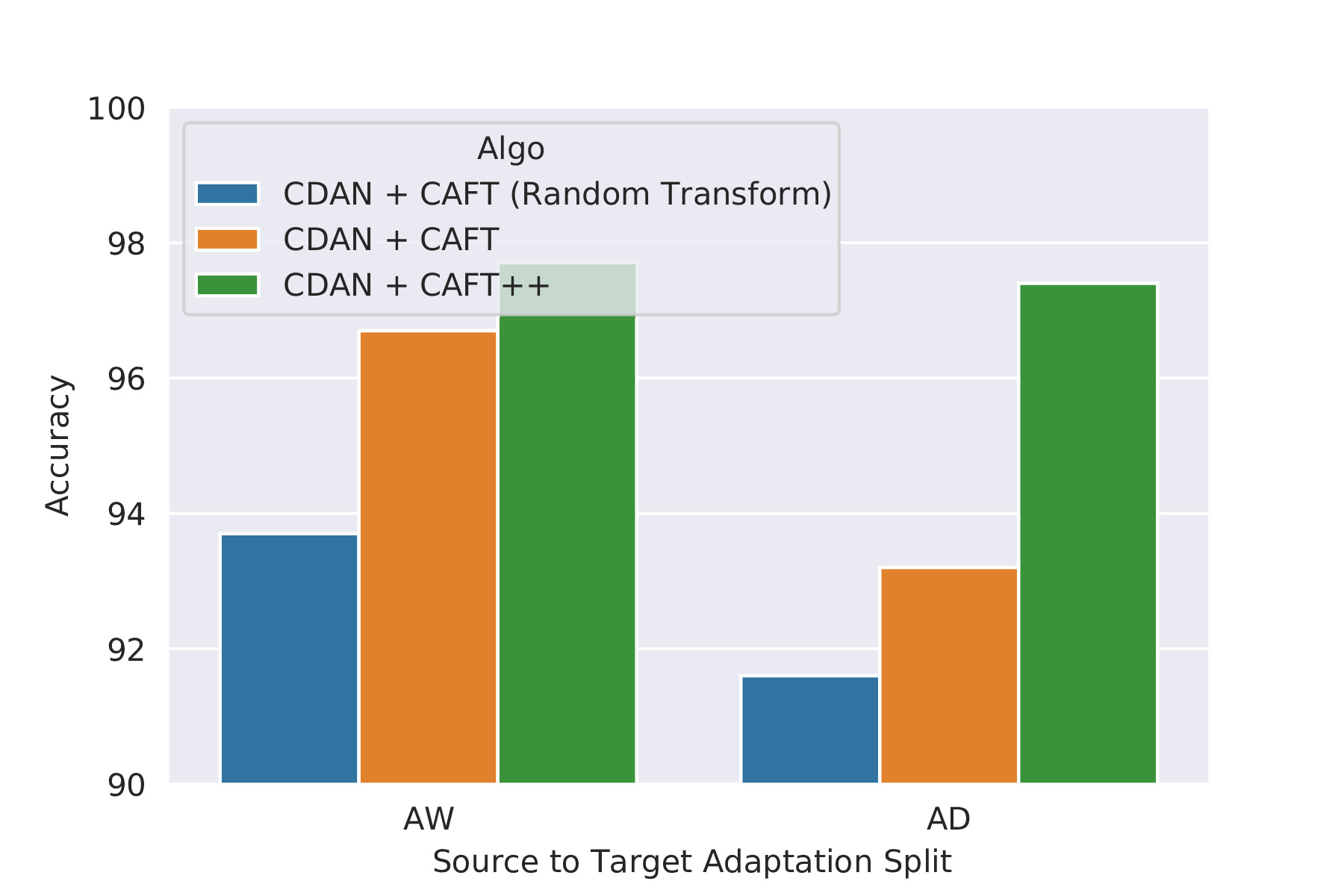}
        
    \end{minipage}%
    \begin{minipage}{0.5\textwidth}
        \centering
        \includegraphics[width=1\linewidth]{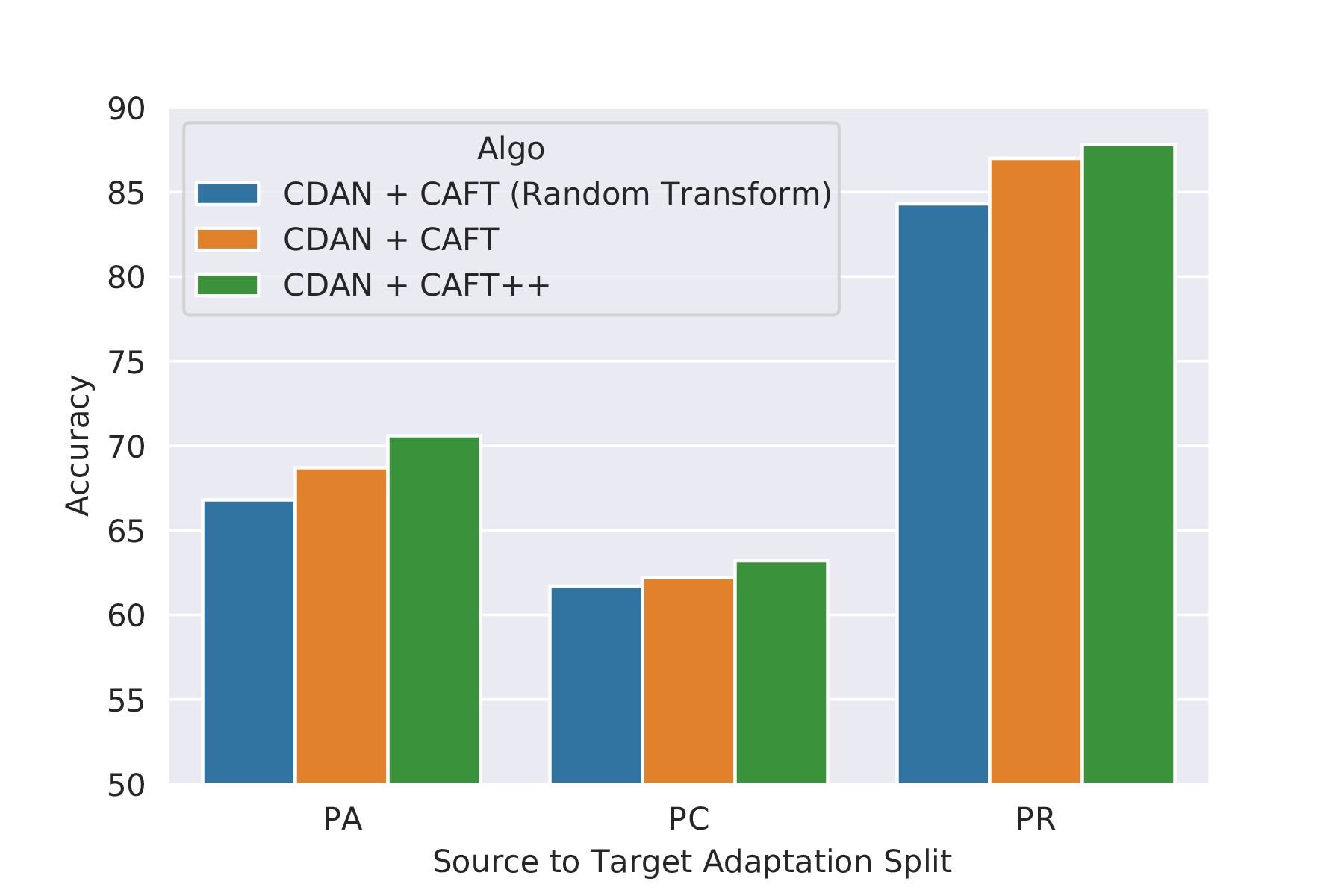}
        
    \end{minipage}%
    \vspace{1mm}
    \caption{The accuracy for domain adaptation task CDAN+CAFT++ against the CDAN+CAFT and CDAN+CAFT with Random Transform for CDAN \cite{long2018conditional}. Left: target adaptation performance from Amazon to Webcam and Amazon to Dslr indicates that CAFT classaware transformation is better than random transformation and CAFT++ provides further gain due to improved design. Right: We observe similar trend across three settings when we perform target domain adaptation from Product to Art, Clipart and Real World respectively.}
    \label{fig:random_caft_cpp}
\end{figure*}

By moving the low-frequency component from the target to the source, FDA \cite{yang2020fda} attempts to close the domain gap. This is because the domain's style may very well be derived from the low-frequency component. The primary issue with FDA \cite{yang2020fda} is the target sample selection. It chooses a random image from from the target domain for style transfer, assuming that style is distributed uniformly throughout the domain, i.e. $\mathcal{Z} \sim \mathcal{U}(a,b)$ where $\mathcal{Z}$ is a random variable representing style sampled uniformly. In many practical situations, the uniform distribution hypothesis for style does not hold true i.e $\mathcal{Z} \not\sim \mathcal{U}(a,b)$, which accounts for inter-class variations. It is fair to assume, however, that the intra-class style for a given domain may be represented with a uniform distribution i.e $\mathcal{Z}_i \sim \mathcal{U}\{a_i,b_i\}$ for $i = 1$ to $K$ where $K$ is the total number of classes for any domain $\mathcal{D}$. For instance, Most of the time, the fish will be present with water. As a result, tackling the classification problem using the FDA \cite{yang2020fda} without taking inter-class information into consideration is not recommended. We can observe from Figure  \ref{fig:acc_curve} and Figure  \ref{fig:loss_curve} that the low frequency component swapping in class-aware manner results in improved accuracy and faster convergence compared to randomly selecting any target sample. We observe a similar trend in target domain accuracy in Figure \ref{fig:random_caft_cpp} when we apply random transformation (CDAN + CAFT(Random Transfrom)), pseudo label based transformation (CDAN + CAFT) and pseudo label using ADT2P (CDAN + CAFT++).  Therefore, selecting target samples in class aware manner is important for improved performance.

We obtain the pseudo label for the target samples from Stage-2. The target samples are stored as key-value pairs, with key corresponding to pseudo class labels. Based on the true class label of the source sample, we randomly select target images from the dictionary i.e $\mathcal{T}_{b} \sim {\mathcal{T}_{pl}\{X_T, Y_{T}^{pl}\}}$ such that $Y_S$ is equal to $Y_{T}^{pl}$ for each sample in the batch of source dataset where $\mathcal{T}_{b}$ is batch of target domain samples, $Y_{T}^{pl}$ is target pseudo label and, $Y_S$ is source true label. During the transform, it assures that target is represented in a class-aware manner. In case some of the pseudo target class label are not present in the dictionary, we randomly select any target sample and continue with the target adaptation. During the initial training phase, we may have a situation where we do not have representation of a particular target class. As the training progresses, we may have representation from almost all the pseudo target classes due to ongoing target adaptation. In CAFT++, we empirically observe that we have at least a few correct images from each class from the start of training. This is because, in CAFT++, we use pre-trained domain adaptation weights for initialization of training. For the current batch, we calculate the Fast Fourier transform (FFT) for the source and target samples across each channel. The FFT method is a fast implementation of the Discrete Fourier Transform (DFT), whose expression is shown in equation (\ref{eq:eq1}) and an inverse representation shown in equation (\ref{eq:f_inverse}).
\begin{equation} \label{eq:eq1}
\begin{split}
\small{\mathcal{F} = \frac{1}{MN}\sum_{m=0}^{M-1}\sum_{n=0}^{N-1} f[n,m]\exp\{-j2\pi(\frac{ kn}{N} + \frac{lm}{M})\}}
\end{split}
\end{equation}
\begin{equation} \label{eq:f_inverse}
\begin{split}
\small{\mathcal{F}^{-1} = \frac{1}{KL}\sum_{k=0}^{K-1}\sum_{l=0}^{L-1} F[k,l]\exp\{j2\pi(\frac{ kn}{K} + \frac{lm}{L})\}}
\end{split}
\end{equation}
Where $M ~\text{and}  ~N$ are number of rows and columns in the 2-D image. $f[n,m]$ is the pixel value at $m^{th}$ row and $n^{th}$ column index. Similarly, $F[k,l]$ is the frequency value at $k^{th}$ row and $l^{th}$ column index. $\mathcal{F}$ and $\mathcal{F}^{-1}$ are Fourier and inverse Fourier transform respectively.
We need to transfer the target's style to the source, which is present in the form of a low-frequency magnitude component after we compute the FFT for both source and target samples. As demonstrated in Figure  \ref{fig:architecture}(c), we substitute the source's low-frequency magnitude component with that of the target. we refer it as Frequency Transformed source $\hat{\mathcal{S}}$. In image space, it assures a reduced domain shift between Frequency Transformed source $\hat{\mathcal{S}}$ and target domain $\mathcal{T}$.
We summarize the overall frequency transformation procedure in equation (\ref{eq:eq2}).
\begin{equation}\label{eq:eq2}
\hat{\mathcal{S}} = \mathcal{F}^{-1} [ \tilde{f}\{\mathcal{F(\mathcal{S})}, \mathcal{F(\mathcal{T})}\} ]
\end{equation}
Where $\mathcal{F}$ and $\mathcal{F}^{-1}$ are Fourier transform and inverse Fourier transform respectively. Function $\tilde{f}$ represents the low frequency swapping operation as shown in Figure  \ref{fig:architecture}(c). We name this process as \textbf{C}lass \textbf{A}ware \textbf{F}requency \textbf{T}ransformation (CAFT). Please refer to algorithm \ref{alg:algorithm} for complete implementation detail.

\subsection{Target Adaptation and Fine-tuning}
We have access to modified source samples $\hat{\mathcal{S}}$, as well as original source samples $\mathcal{S}$ and target samples $\mathcal{T}$, at the start of Stage-4. The modified source images are represented by the center circle in the bottom row of Figure \ref{fig:overview_img}. In comparison to the original source samples $\mathcal{S}$ and target sample $\mathcal{T}$, we see a finite decrease in the domain shift between converted source sample $\hat{\mathcal{S}}$ and target sample $\mathcal{T}$. 
When we use frequency transformed source samples $\hat{\mathcal{S}}$ during domain adaptation, we observe that the feature representation loses their class-discriminative ability but improves the transferability. Please refer to ablation plot in Figure  \ref{fig:abla_without_source}, where we observe the reduced target accuracy when we only use the transformed source. It happens because the replacement of low-frequency magnitude from the target samples affects both background as well as the foreground of the source image which resulting into artefacts in images created after taking the inverse Fourier transformation. 
The trained model's classification performance on source and target images is impacted by these artifacts. If we wish to maintain non-decreasing model performance, we must guarantee that the class-discriminative features do not deteriorate. To tackle this problem, we integrate both original and modified source samples during adaptation. Finally, we have $\overline{S} = \mathcal{S} ~\cup~ \hat{\mathcal{S}} $ as our new source domain sample, and we perform adaptation between $\overline{S}$ and $\mathcal{T}$ using existing domain adaptation algorithms.
\noindent\textbf{Fine-tuning:} We get very confident clean pseudo labels as a result of our proposed pseudo label filtering scheme. Since the noise content is very less, It provides an opportunity to improve target performance by applying fine-tuning. We use pseudo label based cross entropy loss Eq. \ref{eq:pl_ce} only on the clean samples from stage-2 for fine-tuning the adapted model.
\noindent Pseudo labels based cross entropy loss is defined in equation \ref{eq:pl_ce}. 
\begin{equation}\label{eq:pl_ce}
\resizebox{\columnwidth}{!}{$
\mathcal{L}_{ce}^{pl} =\frac{-1}{N_{cl}}
\sum_{i=1}^{N_{cl}}\sum_{k=1}^{K}
\{\mathbbm{1}{[k={y}_{i}^{pl}]}\log(\delta_{k}(g(x_{i};\theta_{s},\phi_{s})))\}
$}
\end{equation}
\noindent Where $N_{cl}$ is the total number of clean samples and  $y_{i}^{pl}$ is pseudo label for the $i^{th}$ clean sample, K is the total number of classes and $\delta_k$ is softmax output for $k^{th}$ class.

\section{Experiments}
\label{sec:exp}

\begin{figure*}[t]
    \centering
    \begin{minipage}{0.5\textwidth}
        \centering
        \includegraphics[width=1\linewidth]{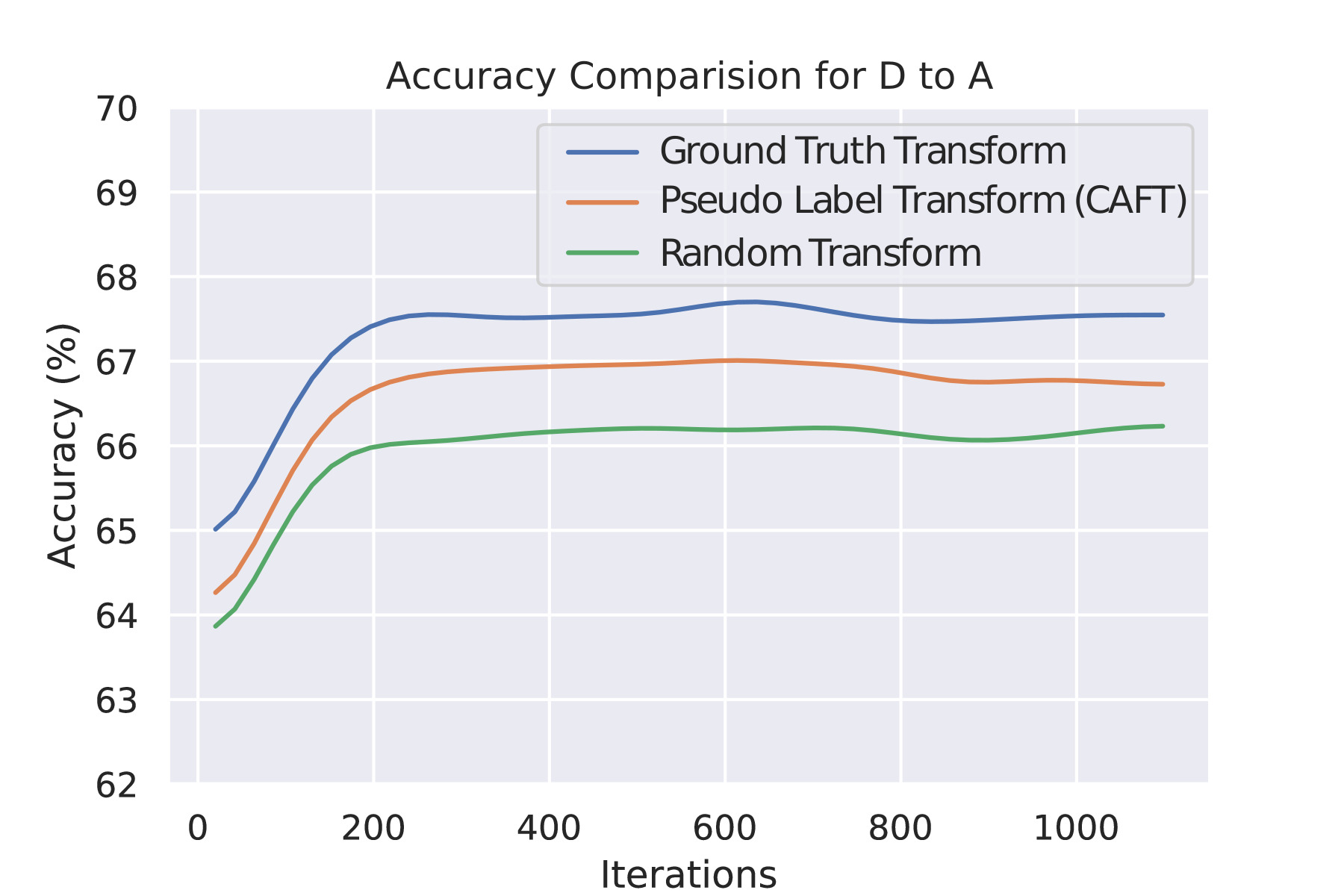}
    \end{minipage}%
    \begin{minipage}{0.5\textwidth}
        \centering
        \includegraphics[width=1\linewidth]{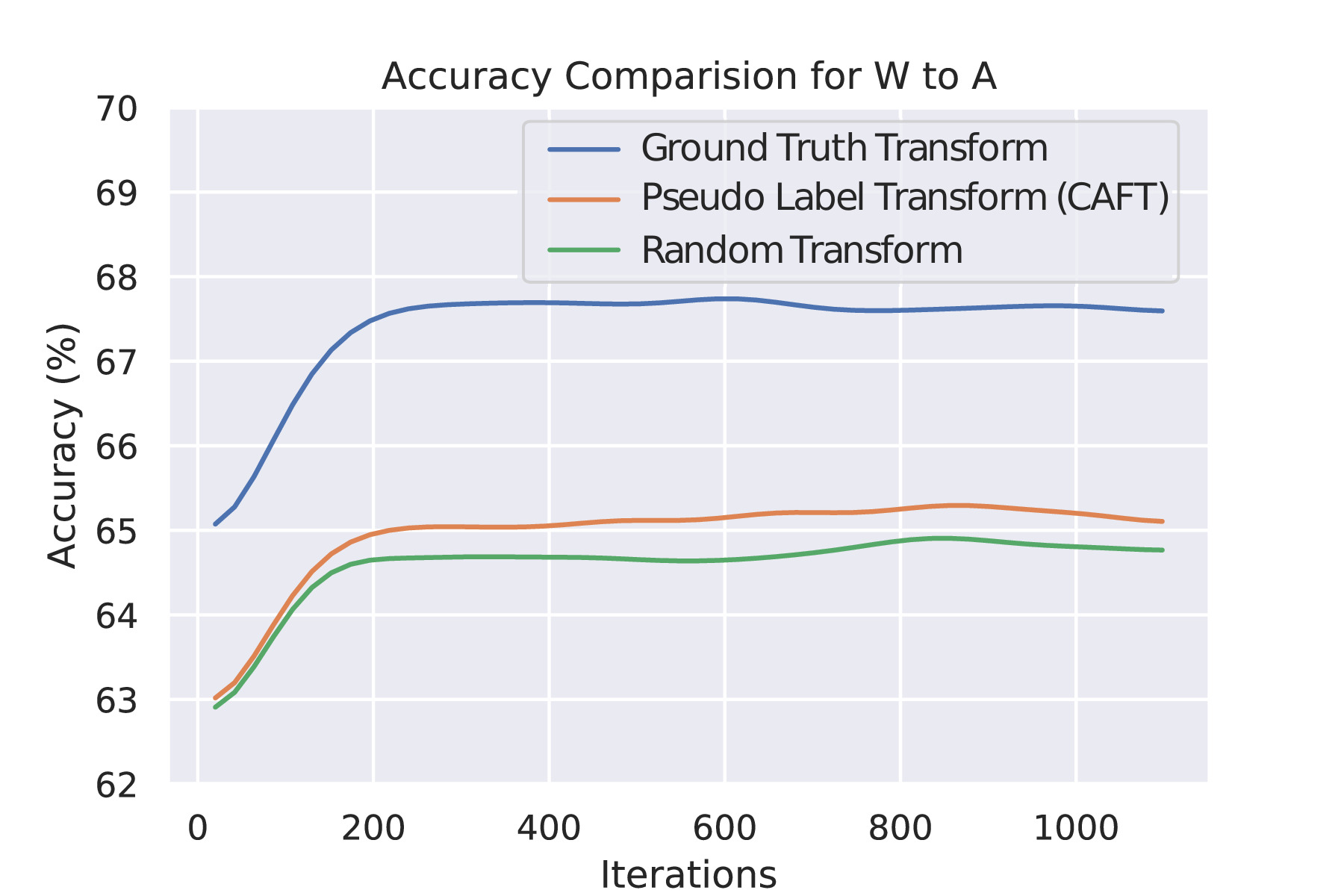}
    \end{minipage}
    \vspace{1mm}
    \caption{The accuracy plots for ideal case ground truth based target transformation against the pseudo label based transformation (CAFT) and random transformation method for DeepCoral \cite{sun2016deep}. The superior performance of pseudo label based transformation (CAFT) against the random transformation is evident from the graphs. Due to noisy predictions in the pseudo labels, model performance stays between ground truth based target transformation and random transformation. In ground truth based target transformation case, we assume to have access to the ground truth. In random transformation, we randomly select target samples for CAFT.}
    \label{fig:acc_curve}
\end{figure*}


The Proposed approach is model agnostic and can be plugged into any adaptation method for improving its performance. We evaluate the proposed method on popular DSAN \cite{zhu2020deep}, DANN \cite{ganin2015unsupervised} and DeepCoral \cite{sun2016deep} using Office-31, Office-Home and large-scale VisDA datasets. These are widely reported domain adaptation benchmarks for classification.

\subsection{Dataset}

\textbf{Office-31 \cite{saenko2010adapting} :} It is divided into three domains, each having 31 classes. There are a total of 4110 picture samples in this collection, and the domains are Amazon (\textit{A}), Webcam (\textit{W}) and DSLR (\textit{D}). The Amazon dataset has a white background and was acquired from amazon.com. Images from a webcam and a DSLR camera were taken in an office setting. For Webcam and DSLR, the resolution disparity works as a domain gap. Images taken with a DSLR have a high quality, whereas those taken with a webcam have a low resolution. There are 6 possible adaptation setting for this dataset which are $A \rightarrow W$, $A \rightarrow D$, $D \rightarrow A$, $D \rightarrow W$, $W \rightarrow A$ and $W \rightarrow D$.

\noindent\textbf{Office-Home \cite{venkateswara2017deep} :} It is divided into four domains, each with 65 classes. There are 15588 image samples in all. 4 domains are Art(\textit{A}), Product(\textit{P}), Real World (\textit{R}) and Clipart (\textit{C}). This is one of the most often used benchmarking datasets for domain adaptation. It also aids in evaluating the scalability of our suggested solution because it contains higher number of classes.

\noindent\textbf{VisDA \cite{peng2017visda} :} VisDA is a large scale dataset that has 12 classes and two domains in the form of a synthetic image and a real image dataset. The synthetic dataset is the source domain, containing 152397 image samples, while the real image dataset is the target domain, containing 55388 image samples. Domain adaptation happens between synthetic to real dataset therefore, the domain shift is higher compared to other datasets.

\subsection{Experiment Setup Details}
\begin{table*}[ht!]
\centering
\caption{Accuracy (\%) on Office-31 (ResNet50)}

\label{tab:Off31}
\resizebox{0.7\textwidth}{!}{\begin{tabular}{@{}c|ccccccc@{}}
\toprule
Method & A$\rightarrow$W & D$\rightarrow$W & W$\rightarrow$D & A$\rightarrow$D & D$\rightarrow$A & W$\rightarrow$A & Avg \\
\midrule
ResNet & 68.4 & 96.7 & 99.3 & 68.9 & 62.5 & 60.7 & 76.1\\
DDC~\cite{tzeng2014deep} & 75.8 & 95.0 & 98.2 & 77.5 & 67.4 & 64.0 & 79.7\\
DAN~\cite{long2015learning} & 83.8 & 96.8 & 99.5 & 78.4 & 66.7 & 62.7 & 81.3\\
ADDA~\cite{tzeng2017adversarial} & 86.2 & 96.2 & 98.4 & 77.8 & 69.5 & 68.9 & 82.9\\
JAN~\cite{long2017deep} & 85.4 & 97.4 & 99.8 & 84.7 & 68.6 & \textbf{70.0} & 84.3\\
MADA~\cite{pei2018multi} & 90.0 & 97.4 & 99.6 & 87.8 & 70.3 & 66.4 & 85.2\\
CAN~\cite{zhang2018collaborative} & 81.5 & 98.2 & 99.7 & 85.5 & 65.9 & 63.4 & 82.4\\
iCAN~\cite{zhang2018collaborative} & \textbf{92.5} & \textbf{98.8} & \textbf{100.0} & \textbf{90.1} & \textbf{72.1} & {69.9} & \textbf{87.2}\\

\midrule

DeepCoral \cite{sun2016deep} & 77.7 & 97.6 & 99.7 & 81.1 & 64.6 & 64.0 & 80.8 \\
DeepCoral + CAFT& 79.4  & \textbf{97.9} & \textbf{100.0} & 81.5 & \textbf{66.6} & 65.7 & 81.8 \\
DeepCoral + CAFT++ & \textbf{82.1}  & \textbf{97.9} & \textbf{100.0} & \textbf{84.7} & 66.1 & \textbf{66.7} & \textbf{82.9} \\

\midrule

DANN \cite{ganin2015unsupervised} & {91.4} & {97.9} & {\textbf{100.0}} & {83.6} & 73.3 & 70.4 & 86.1\\
DANN + FGDA \cite{gao2021gradient} &  \textbf{92.6} & \textbf{99.1} & \textbf{100.0} & \textbf{94.2} & 73.9 & 73.7 & \textbf{88.9} \\
DANN + CAFT & 84.2 & 97.6 & 99.6 & 81.3 & 70.2 & 72.9 & 84.3\\
DANN + CAFT++ & 90.4 & 97.6 & \textbf{100.0} & 83.5 & \textbf{74.8} & \textbf{74.4} & 86.7\\
\midrule
DSAN \cite{zhu2020deep} & 93.6 & 98.3 & \textbf{100.0} & \textbf{90.2} & 73.5 & 74.8 & 88.4\\
DSAN + CAFT & 92.0  & 98.0 & \textbf{100.0} & 88.0 & 75.0 & 75.0 & 88.0 \\
DSAN + CAFT++ & \textbf{93.8}  & \textbf{98.7} & \textbf{100.0} & 89.6 & \textbf{78.9} & \textbf{77.3} & \textbf{89.7} \\

\midrule
MCD \cite{saito2018maximum} & 91.1 & 98.2 & \textbf{100.0} & 88.4 & 68.2 & 67.7 & 85.6\\
MCD + CAFT & 84.7 & 98.2 & 100.0 & 84.9 & 67.5 & 66.5 & 83.6\\
MCD + CAFT++ & \textbf{91.6} & \textbf{98.4} & \textbf{100.0} & \textbf{88.6} & \textbf{69.1} & \textbf{68.3} & \textbf{86.0}\\

\midrule
MCC \cite{jin2020minimum} & 94.6 & 98.2 & 99.6 & 93.8 & 74.7 & 74.2 & 89.2\\
MCC + SCDA \cite{li2021semantic} & 93.7 & \textbf{98.6} & \textbf{100.0} & \textbf{96.4} & \textbf{76.5} & \textbf{76.0} & \textbf{90.2}\\
MCC + CAFT & 92.1 & 92.9 & 99.8 & 92.8 & 74.1 & {75.8} & 88.9\\
MCC + CAFT++ & \textbf{95.8} & {98.5} & {99.8} & {95.0} & {75.9} & 75.2 & {90.0}\\

\midrule
MDD \cite{zhang2019bridging} & 94.2&98.7&\textbf{100.0}&94.6&75.6&72.7&89.3 \\
MDD + SCDA \cite{li2021semantic} & 95.3 & 99.0 & \textbf{100.0} & 95.4 & 77.2 & 75.9 & 90.5\\
MDD + FGDA \cite{gao2021gradient} & 95.1 & 98.7 & \textbf{100.0} & 95.4 & 78.1 & 76.5 & 90.6\\
MDD + CAFT & 96.4&	\textbf{99.6}&	\textbf{100.0} &	94.8&	80.7&	\textbf{80.5}&	92.0\\
MDD + CAFT++ & \textbf{96.9}&99.2&\textbf{100.0}&\textbf{96.0}&\textbf{80.9}&79.6&\textbf{92.1}\\

\midrule
CDAN \cite{long2018conditional} & 95 & 98.6 & \textbf{100.0} & 95.8 & 73.8 & 73.3 & 89.4 \\
CDAN + SCDA \cite{li2021semantic} & 94.7&98.7&\textbf{100.0}&95.4&77.1&76&90.3\\
CDAN + FGDA \cite{gao2021gradient} & 92.6&98.7&\textbf{100.0}&95&74.7&74.4&89.2\\
CDAN + RADA \cite{jin2021re} & 96.2&99.3&\textbf{100.0}&96.1&77.5&77.4&91.1\\
CDAN + ATDOC-NC \cite{liang2021domain} &93.6&99.1&\textbf{100.0}&96.3&74.3&75.4&89.8 \\
CDAN + ATDOC-NA \cite{liang2021domain} &94.6&98.1&99.7&95.4&75.5&77&90.1\\
CDAN + CAFT & 96.7&	99.5&	\textbf{100.0}&	93.2&	\textbf{80.7} &	80 &	91.7 \\
CDAN + CAFT++ & \textbf{97.7}&\textbf{99.5}&\textbf{100.0}&\textbf{97.4}& 79.4 &\textbf{81.1}&\textbf{92.5}\\
\bottomrule
\end{tabular}}
\end{table*}

In our experiments, we use the ResNet-50 \cite{he2016deep} backbone as a feature extractor for Office-Home and Office-31 dataset and ResNet-101 for VisDA dataset. The weights were updated using a minibatch stochastic gradient descent algorithm with a momentum of 0.9 and a decreasing learning rate as a function of epoch, as stated in equation (\ref{eq:eq5}) following the implementation details of  DANN \cite{ganin2015unsupervised} :
\begin{equation} \label{eq:eq5}
    \mu_p = \frac{\mu_0}{(1 + 10p)^\beta}
\end{equation}
\begin{equation} \label{eq:eq6}
    p = \frac{epoch}{total ~number~ of~ epochs} 
\end{equation}
\noindent where $p \in [0,1]$ and $\mu_0$ (initial learning rate) as $0.01$.

The side of square window size for selecting low-frequency component is $ 2L*\min(Height, Width)$ and $L$ was chosen to be 0.04 for DSAN, DANN and DeepCoral. Batch size was 24 for Office-31 and Office-Home and 48 in case of VisDA. We use model accuracy as a metric for evaluating the proposed approach.

\subsection{Results}
We represent results for different methods within a block in different tables. We highlight the best result within the block in bold. We discuss experiment results for each methods in sub-sequent paragraphs.

\begin{table*}[h!]
\centering
\caption{Accuracy (\%) on Office-Home (ResNet50)} 
\label{tab:OffHome}
\resizebox{\textwidth}{!}{\begin{tabular}{c|ccccccccccccc}                           
\toprule
Method & A$\rightarrow$C & A$\rightarrow$P & A$\rightarrow$R & C$\rightarrow$A & C$\rightarrow$P & C$\rightarrow$R & P$\rightarrow$A & P$\rightarrow$C & P$\rightarrow$R & R$\rightarrow$A & R$\rightarrow$C & R$\rightarrow$P & Avg \\
\midrule
ResNet & 34.9 & 50.0 & 58.0 & 37.4 & 41.9 & 46.2 & 38.5 & 31.2 & 60.4 & 53.9 & 41.2 & 59.9 & 46.1\\
DAN~\cite{long2015learning} & 43.6 & 57.0 & 67.9 & 45.8 & 56.5 & 60.4 & 44.0 & \textbf{43.6} & 67.7 & 63.1 & 51.5 & 74.3 & 56.3\\
JAN~\cite{long2017deep} & \textbf{45.9} & \textbf{61.2} & \textbf{68.9 } & \textbf{50.4 }& \textbf{59.7} &\textbf{ 61.0 }& \textbf{45.8} & 43.4 & \textbf{70.3 }& \textbf{63.9} & \textbf{52.4} & \textbf{76.8} & \textbf{58.3}\\

\midrule
DeepCoral \cite{sun2016deep} & 52.3 & 69.0 & 75.9 & 56.7 & 66.1 & 68.7 &
55.8 & 46.9 & 75.3 & 67.0 & 53.0 & 80.0 & 63.9\\
DeepCoral + CAFT & 53.5 & 69.4 & \textbf{76.4} & \textbf{58.9} & 67.1 & 69.6 & 56.3 & 48.4 & 76.0 & 67.0 & 55.0 & 80.2 & 64.8\\
DeepCoral + CAFT++ & \textbf{54.5} & \textbf{70.5} & 76.3 & 58.8 & \textbf{67.8} & \textbf{70.0} & 
\textbf{59.4} & \textbf{48.8} & \textbf{76.2} & \textbf{67.7} & \textbf{55.9} & \textbf{80.6} & \textbf{65.5}\\ 
\midrule
DANN \cite{ganin2015unsupervised} & 53.8 & 62.6 & 74.0 & 55.8 & \textbf{67.3} & 67.3 &
55.8 & \textbf{55.1} & 77.9 & 71.1 & 60.7 & \textbf{81.1} & 65.2\\
DANN + CAFT & 52.4 & 60.8 & 73.9 & 53.9 & 63.9 & 66.3 & 55.1 & 53.8 & 76.1 & 68.9 & 60.1 & 80.7 & 63.8\\
DANN + CAFT++ & \textbf{54.3} & \textbf{63.6} & \textbf{75.0} & \textbf{58.2} & \textbf{67.3} & \textbf{68.8} & \textbf{59.0} & 54.8 & \textbf{78.1} & \textbf{71.4} & \textbf{61.4} & 80.4 & \textbf{66.0}\\
\midrule
DSAN \cite{zhu2020deep} & 54.4 & 70.8 & 75.4 & 60.4 & 67.8 & 68.0 & 62.6 & 55.9 & 78.5 & 73.8 & 60.6 & 83.1 & 67.6\\
DSAN + CAFT~ & 55.2 & 69.8 & 75.0 & 60.0 & 72.0 & 71.0 & 63.3 & 57.3 & 79.1 & 74.1 & 60.6 & 83.0 & 68.4\\
DSAN + CAFT++ & \textbf{60.7} & \textbf{76.0} & \textbf{78.8} &
\textbf{67.1} & \textbf{75.8} & \textbf{74.5} & 
\textbf{66.8} & \textbf{59.3} & \textbf{81.7} & \textbf{74.3} & \textbf{64.6} & \textbf{85.3} & \textbf{72.1}\\
\midrule
MCD \cite{saito2018maximum} & \textbf{51.0} & 72.3 & 77.7 & 61.1 & 68.9 & 68.7 & 59.4 & 50.7 & 78.0 & 73.9 & \textbf{58.1} & 81.3 & 66.8\\
MCD + CAFT~ & 47.3 & 66.3 & 76.5 & 61.7 & 67.4 & 69.4 & 60.1 & \textbf{51.2} & 76.7 & 73.4 & 57.1 & 81.3 & 65.7\\
MCD + CAFT++ & 50.7 & \textbf{73.5} & \textbf{78.3} & \textbf{62.5} & \textbf{69.6} & \textbf{70.6} & \textbf{60.5} & 50.3 & \textbf{79.5} & \textbf{75.0} & 57.4 & \textbf{82.5} & \textbf{67.5}\\
\midrule
MCC \cite{jin2020minimum} & 58.0 & 79.1 & 82.9 & 66.8 & 78.4 & 78.4 & 66.7 & \textbf{54.8} & 81.7 & 74.4 & 61.0 & \textbf{85.6} & 72.3\\
MCC + SCDA \cite{li2021semantic}~ & 57.1 & 79.1 & 82.7 & \textbf{67.7} & 75.3 & 77.6 & 66.3 & 52.5 & 81.9 & \textbf{74.9} & 60.1 & 85.0 & 71.7\\
MCC + CAFT~ & 56.3 & 76.3 & 77.7 & 65.1 & 74.4 & 74.4 & 64.1 & 51.6 & 78.0 & 71.7 & 54.9 & 83.5 & 69.0\\
MCC + CAFT++ & \textbf{59.0} & \textbf{80.2} &\textbf{ 83.3} & {67.2} & \textbf{78.5} & \textbf{79.5} & \textbf{67.7} & 54.5 & \textbf{82.2} & \textbf{74.9} &\textbf{61.3} & 85.4 & \textbf{72.8}\\
\midrule
CDAN \cite{long2018conditional} & 53.4&71.6&77&61.5&71.9&72.2&61.8&54.8&80.8&74.7&60.6&80.9&68.4\\
CDAN  + SCDA \cite{li2021semantic}~ & 57.1 & 75.9 & 79.9 & 66.2 & 76.7 & 75.2 & 65.3 & 55.6 & 81.9 & 74.7 & 62.6 & 84.5 & 71.3\\
CDAN + RADA \cite{jin2021re}~ & 56.5 & 76.5 & 79.5 & 68.8 & 76.9 & 78.1 & 66.7 & 54.1 & 81.0 & 75.1 & 58.2 & 85.1 & 71.4\\
CDAN + ATDOC-NC \cite{liang2021domain} & 55.9&	76.3&	80.3&	63.8&	75.7&	76.4&	63.9&	53.7&	81.7&	71.6&	57.7&	83.3&	70\\
CDAN + ATDOC-NA \cite{liang2021domain} & 60.2&	77.8&	82.2&	68.5&	78.6&	77.9&	68.4&	58.4&	83.1&	74.8&	61.5&	87.2&	73.2\\
CDAN + CAFT  & \textbf{62.6}&	78.1&	84.7&	68.9&	\textbf{79.7}&	\textbf{79.9}&	68.7&	62.2&	87.0 &	80.5&	68.9&	89.5&	75.9\\
CDAN + CAFT++  & \textbf{62.6} & \textbf{78.8} & \textbf{85.1} & \textbf{70.4} & 79.2 & 79.6 & \textbf{70.6} & \textbf{63.2} & \textbf{87.8} & \textbf{81.6} & \textbf{69.1} & \textbf{89.7} &\textbf{ 76.5}\\
\midrule
MDD \cite{zhang2019bridging} & 56.1&	75.4&	79.7&	63.4&	72.6&	74.1&	61.8&	54.3&	80.1&	73.5&	60.7&	84.4&	69.7\\
MDD + SCDA \cite{li2021semantic}~ & 58.9 & 77.2 & 81.0 & 66.6 & 75.5 & 75.9 & 64.1 & 56.3 & 82.2 & 73.3 & 61.5 & 84.3 & 71.4\\
MDD + FGDA \cite{gao2021gradient} & 57.1 & 77.5 & 81.0 & 68.4 & 77.2 & 75.9 & 65.8 & 55.8 & 81.0 & 74.3 & 60.5 & 83.6 & 71.5\\
MDD + CAFT  & 66.0&	\textbf{83.5}&	86.9&	\textbf{73.3}&	\textbf{80.6}&	\textbf{82.4}&	\textbf{71.9}&	\textbf{65.5}&	87.6&	\textbf{81.1}&	\textbf{71.6}&	90.4&	\textbf{78.4}\\
MDD + CAFT++  & \textbf{66.2} & 83.3 & \textbf{87.7} & 72.4 & 80.4 & \textbf{82.4} & 71.4 & 63.2 & \textbf{87.9} & 80.7 & 71.4 & \textbf{90.8} & 78.2\\
\bottomrule
\end{tabular}}
\end{table*}

\begin{table*}[h!]
\centering
\caption{H-Score (\%) on Office-Home Open set} 
\label{tab:DANN_OffHome_openset}
\resizebox{\textwidth}{!}{\begin{tabular}{c|ccccccccccccc}
\toprule
Method & A$\rightarrow$C & A$\rightarrow$P & A$\rightarrow$R & C$\rightarrow$A & C$\rightarrow$P & C$\rightarrow$R & P$\rightarrow$A & P$\rightarrow$C & P$\rightarrow$R & R$\rightarrow$A & R$\rightarrow$C & R$\rightarrow$P & Avg \\
\midrule
DANN \cite{ganin2015unsupervised} & 56.0 & 66.2 & \textbf{74.3} & 60.3 & 64.5 & \textbf{68.2} & 60.3 & 53.7 & \textbf{70.0} & 65.5 & 59.9 & \textbf{72.4} & 64.3\\
DANN + CAFT & \textbf{58.6} & 67.1 & 73.3 & \textbf{60.7} & 64.3 & 67.2 & 60.9 & \textbf{54.9} & 68.4 & 66.1 & \textbf{60.2} & 71.5 & 64.4\\
DANN + CAFT++ & 57.7 & \textbf{67.7} & 73.5 & 60.5 & \textbf{65.3} & 67.4 & \textbf{62.1} & 54.0 & 69.5 & \textbf{66.8} & 59.3 & 71.5 & \textbf{64.6}\\
\midrule
OSBP \cite{saito2018open} & 61.5&71.7&76.2&65.9&69.2&73.1&65.9&59.1&76.4&71.6&61.6&77.7&69.2\\
OSBP + CAFT & \textbf{63.8}&71.6&76.6&64.5&69.5&73.0&\textbf{66.7}&58.2&\textbf{77.0}&72.0&60.9&77.6&69.3\\
OSBP + CAFT++ & \textbf{63.8}&\textbf{72.0}&\textbf{76.7}&\textbf{66.6}&\textbf{69.9}&\textbf{73.9}&66.4&\textbf{59.2}&76.6&\textbf{72.2}&\textbf{63.2}&\textbf{78.4}&\textbf{69.9}\\
\bottomrule
\end{tabular}}
\end{table*}

\begin{table*}[h!]
\centering
\caption{Accuracy (\%) on Office-Home Partial set} 
\label{tab:DANN_OffHome_partialset}
\resizebox{\textwidth}{!}{\begin{tabular}{c|ccccccccccccc}
\toprule
Method & A$\rightarrow$C & A$\rightarrow$P & A$\rightarrow$R & C$\rightarrow$A & C$\rightarrow$P & C$\rightarrow$R & P$\rightarrow$A & P$\rightarrow$C & P$\rightarrow$R & R$\rightarrow$A & R$\rightarrow$C & R$\rightarrow$P & Avg \\
\midrule
DANN \cite{ganin2015unsupervised} & 46.0 & 61.6 & \textbf{76.0} & 45.5 & 47.2 & 54.7 & 52.9 & 39.1 & 71.6 & 65.5 & 45.8 & \textbf{72.9} & 56.6\\
DANN + CAFT & \textbf{54.2} & \textbf{71.2} & 45.2 & \textbf{57.1} & \textbf{50.0} & 47.2 & \textbf{68.7} & \textbf{47.9} & 60.8 & 66.9 & 40.5 & 50.7 & 55.0\\
DANN + CAFT++ & 49.4 & 61.5 & 75.6 & 47.1 & 46.0 & \textbf{57.2} & 59.4 & 44.6 & \textbf{72.2} & \textbf{67.0} & \textbf{48.6} & 71.7 & \textbf{58.4}\\
\midrule
PADA \cite{cao2018partial2} & 49.3 & 62.2 & 81.0 & 51.1 & 50.8 & 63.1 & 64.8 & 38.3 & 81.8 & 76.3 & 44.7 & 82.6 & 62.2\\
PADA + CAFT & 51.8&	\textbf{67.5}&	81.6&	\textbf{57.3}&	\textbf{54.3}&	63.2&	66.8&	\textbf{39.8}&	81.8&	\textbf{77.5}&	\textbf{50.6}&	82.1&	\textbf{64.5}\\
PADA + CAFT++ & \textbf{52.2} & 65.4 & \textbf{81.8} & 53.7 & 53.6 &\textbf{ 67.3} & \textbf{68.8} & 36.7 & \textbf{81.9} & 77.2& 45.7 & \textbf{82.7} & 63.9\\

\bottomrule
\end{tabular}}
\end{table*}

\noindent \textbf{DANN:} We show the experiment results for DANN \cite{ganin2015unsupervised} in Table \ref{tab:Off31}, \ref{tab:OffHome} and \ref{tab:VISDA}. The proposed approach improves the existing model accuracy on Office-31 by 0.6\%, Office-Home by 0.8\% and VisDA by 1.8\%. We achieved a commendable performance gain in 10 out of 12 splits for the VisDA dataset.

\begin{table*}[ht]
\centering
\caption{Accuracy (\%) on VisDA (ResNet101) \cite{zhu2020deep}.}
{
\label{tab:VISDA}
\resizebox{\textwidth}{!}{\begin{tabular}{c|ccccccccccccc}
\toprule
Method & aero & bicycle & bus & car & horse & knife & motor & person & plant & skate & train & truck & Avg \\
\midrule
Resnet& 72.3 & 6.1  & 63.4 & \textbf{91.7} & 52.7 & 7.9  & 80.1 & 5.6  & 90.1 & 18.5  & 78.1  & 25.9  & 49.4\\ 
DAN~\cite{long2015learning}   & 68.1 & 15.4 & {76.5} & 87.0 & 71.1 & 48.9 & 82.3 & 51.5 & 88.7 & 33.2  & \textbf{88.9}  & 42.2  & 62.8\\
JAN~\cite{long2017deep}   & 75.7 & 18.7 & 82.3 & 86.3 & 70.2 & 56.9 & 80.5 & 53.8 & \textbf{92.5} & 32.2  & 84.5  &\textbf{ 54.5}  & 65.7\\
MCD~\cite{saito2018maximum}   & \textbf{87.0} & \textbf{60.9} & \textbf{83.7} & 64.0 & \textbf{88.9} & \textbf{79.6} & \textbf{84.7} & \textbf{76.9} & 88.6 & \textbf{40.3 } & 83.0  & 25.8  & \textbf{71.9}\\
\midrule
DSAN \cite{zhu2020deep} & 90.9 & 66.9 & \textbf{75.7} & \textbf{62.4} & 88.9 & \textbf{77.0} & \textbf{93.7} & 75.1 & \textbf{92.8} & 67.6 & 89.1 & 39.4 & 75.1 \\
DSAN + CAFT & 91.5 & 70.1 & 74.9 & 55.1 & 90.2 & 71.0 & 86.9 & \textbf{76.2} & 92.4 & 78.1 & \textbf{91.3} & 45.3 & 76.9 \\
DSAN + CAFT++ & \textbf{94.0} & \textbf{79.4} & 74.5 & 58.0 & \textbf{90.9} & 72.9 & 89.1 & 74.9 & 91.5 & \textbf{87.1} & 90.0 & \textbf{51.9} & \textbf{79.5}\\
\midrule
DANN \cite{ganin2015unsupervised} & 95.0 & \textbf{70.2} & 83.9 & 43.8 & 86.0 & 88.6 & 88.4 & \textbf{77.4} & 87.2 & \textbf{92.4} & \textbf{90.1} & 43.1 & 78.8 \\
DANN + CAFT & 88.4 & 57.3 & 82.8 & \textbf{54.6 }& 77.9 & 91.0 & 83.0 & 74.1 & 82.8 & 68.6 & 80.2 & 39.8 & 73.4 \\
DANN + CAFT++ & \textbf{95.3} & 69.5 & \textbf{84.9} & 50.4 & \textbf{88.4} & \textbf{91.5} & \textbf{92.0} & 76.1 & \textbf{90.8} & 89.8 & 89.9 & \textbf{43.6} & \textbf{80.2} \\
\bottomrule
\end{tabular}}
}
\end{table*}

\noindent\textbf{DSAN:} We show the experiment results for DSAN \cite{zhu2020deep} in Table \ref{tab:Off31} for Office-31, in Table \ref{tab:OffHome} for Office-Home and Table \ref{tab:VISDA} for VisDA. DSAN \cite{zhu2020deep} is one of the state-of-the-art approaches for UDA. The proposed approach improved the accuracy by a significant margin of 3.7\%. We also observed 4.4\% gain VisDA dataset. Since the VisDA dataset size is large, hence 4.4\% gain is non-trivial. In Table. \ref{tab:precision_recall}, we compute precision and recall values for Office-Home dataset. We can observe that overall CAFT++ has higher precision than CAFT. We want to maximize the amount of correctly classified confident samples (true positive) and minimize the amount of wrongly classified confident samples (false positives), even if true positives are less in number. Using this insight, we need to develop a method that has a higher precision value. Our assumption is that the count of true positives will keep increasing, and the count of false positives will keep decreasing as training progresses. The precision of CAFT++ is 3.4\% more than the precision of CAFT in Table. \ref{tab:precision_recall}. The recall does not play enough role in our pseudo labelling approach because we perform style transfer only using the samples from a clean set. Even though the CAFT has a better recall value compared to CAFT++, the final target accuracy obtained after domain adaptation is more in the case of  CAFT++. 

\noindent\textbf{DeepCoral:} Table \ref{tab:Off31} and Table \ref{tab:OffHome} illustrate the experimental findings on DeepCoral \cite{sun2016deep} for Office-31 and Office-Home respectively. Our suggested method enhances the current model's performance over almost all possible Office-31 splits. It results in 1.6\% and 1.1\% absolute increase in average accuracy of Office-Home and Office-31 respectively.

\noindent\textbf{MCC:} We show the result of MCC \cite{jin2020minimum} with CAFT++ for Office-Home and Office-31 in Table \ref{tab:Off31} and \ref{tab:OffHome} respectively. CAFT++ improves the existing performance on Office-Home and on Office-31 by 0.5\% and 0.8\% respectively. Please note that we directly use the parameter given in the original paper and we do not optimize any hyper-parameter for MCC with CAFT++ therefore, results shown in Tables \ref{tab:OffHome} and \ref{tab:Off31} can be improved even further with help of hyperparameter tuning. Further, We also validate that there is considerable gap in model performance for Top-1 and Top-2 accuracy (Table \ref{tab:top1_top2}), which indicate the closeness of top-2 prediction probabilities. ADT2P aims to leverage the same for filtering confusing and confident samples. 

\noindent\textbf{MCD:} We show the result for MCD \cite{saito2018maximum} with CAFT and CAFT++ in Table \ref{tab:Off31} and \ref{tab:OffHome} respectively. CAFT++ improves the existing performance on Office-Home and on Office-31 by 0.7\% and 0.4\% respectively. Please note that we directly use the parameter given in the original paper and we do not optimize any hyper-parameter for MCD with CAFT++ therefore, results shown in Tables \ref{tab:OffHome} and \ref{tab:Off31} can be improved even further with the help of hyperparameter tuning.

\noindent\textbf{CDAN:} We perform experiment by applying CAFT and CAFT++ on CDAN \cite{long2018conditional} and show the results in Table \ref{tab:Off31} for Office-31 and in Table \ref{tab:OffHome} for Office-Home dataset respectively. CAFT++ achieves the best result among recent approaches such as \cite{liang2021domain, li2021semantic, jin2021re, gao2021gradient}.

\noindent\textbf{MDD:} We perform experiment by applying CAFT and CAFT++ on CDAN \cite{zhang2019bridging} and show the results in Table \ref{tab:Off31} for Office-31 and in Table \ref{tab:OffHome} for Office-Home dataset respectively. CAFT++ 

\noindent For all the above experiment tables, one can note that CAFT \cite{Kumar_2021_ICCV} was able to improve upon existing Domain Adaptation algorithm. The proposed CAFT++ is then further able to improve the accuracy of CAFT. This shows the effectiveness of simple new components introduced in CAFT in order to develop CAFT++ algorithm.

\noindent\textbf{CAFT++ Versus Generative Methods:} We compared our frequency based augmentation against popular approach generative methods for style transfer \cite{huang2017arbitrary, zhu2017unpaired}. We replaced our frequency swapping augmentation in CAFT++ with methods like AdaIN \cite{huang2017arbitrary} and CycleGAN \cite{zhu2017unpaired}. We first train these generative models and then obtain the transformed source image using this trained model through inference. From Table \ref{tab:GAN_compare}, we can see that both CycleGAN and AdaIN were able to outperform our proposed frequency swapping method by $0.9\%$ and $0.4\%$ respectively. Even though CycleGAN and AdaIN based style transfer methods produces slightly better results, training and inference time of generative models is very high compared to our frequency swapping method. We also believe that the performance of generative model will further improve if we could train it for longer time.

\noindent\textbf{UDA with Label Shift:} To evaluate the generalization ability of our proposed approach, we run experiments for label shift setup where there can be unknown or private classes in either source or target. We use the DANN architecture with ResNet50 backbone for Open set and Partial set experiments and report the results in Table. \ref{tab:DANN_OffHome_openset} and Table. \ref{tab:DANN_OffHome_partialset} respectively. For Open set, we report the H-Score which is computed 
by taking the harmonic mean of known class accuracy and unknown class accuracy. We can infer that for both partial set and open set UDA setting, CAFT++ generalize well and shows absolute improvement of $1.8\%$ and $0.3\%$ respectively. Further, We apply CAFT and CAFT++ on PADA \cite{cao2018partial2} and performed experiments on the Office-Home dataset. We observe performnace gain across 6 out of 12 tasks for CAFT over CAFT++ and remaining 6 out of 12 tasks for CAFT++ over CAFT. We see large gains in performance for PADA with CAFT and PADA with CAFT++ compared to the base method (only PADA).  We also combine CAFT and CAFT++ with OSBP \cite{saito2018open} and observe an average performance improvement of 0.7\% for OBSP with CAFT++ compared to the base method (only OBSP) on Office-Home dataset. Moreover, when combined with OBSP, CAFT++ outperforms CAFT across 9 out of 12 tasks on the same dataset with an absolute gain of 0.6\% on average performance. This analysis reflects that the effectiveness of CAFT++ is not just limited to standard UDA settings but can also be used for other UDA settings, such as UDA with label shift (Partial and Open-set UDA).

\begin{table*}[h!]
\centering
\caption{Accuracy (\%) comparison for Office-Home with DANN  with different augmentation methods.} 
\label{tab:GAN_compare}

\resizebox{\textwidth}{!}{\begin{tabular}{l|ccccccccccccc}
\toprule
Method & A$\rightarrow$C & A$\rightarrow$P & A$\rightarrow$R & C$\rightarrow$A & C$\rightarrow$P & C$\rightarrow$R & P$\rightarrow$A & P$\rightarrow$C & P$\rightarrow$R & R$\rightarrow$A & R$\rightarrow$C & R$\rightarrow$P & Avg \\
\midrule
CAFT & 52.4 & 60.8 & 73.9 & 53.9 & 63.9 & 66.3 & 55.1 & 53.8 & 76.1 & 68.9 & 60.1 & 80.7 & 63.8 \\
CAFT++  & 54.3 & 63.6 & 75.0 & 58.2 & 67.3 & 68.8 & 59.0 & \textbf{54.8} & \textbf{78.1} & \textbf{71.4} & 61.4 & 80.4 & 66.0 \\
CAFT++ with AdaIN & \textbf{54.8} & 65.1 & 75.7 & 58.1 & \textbf{68.1} & 69.4 & 60.8 & 54.2 & 77.7 & 71.0 & \textbf{61.5} & 80.6 & 66.4 \\
CAFT++ with CycleGAN & 54.3 & \textbf{66.8} & \textbf{76.7} & \textbf{58.6} & 67.8 &\textbf{ 69.8} & \textbf{62.8} & 54.0 & 79.0 & \textbf{71.4} & 61.1 & \textbf{80.9} & \textbf{66.9} \\
\bottomrule
\end{tabular}}
\end{table*}

\section{Ablation Study}
\label{sec:ana}

In this section, we will empirically analyse the proposed approach of class aware Frequency transformation for reducing domain shift. We will also examine the accuracy and convergence of the proposed approach along with comparing it with generative model for style transfer.

\subsection{Analysis of Class Aware Transformation}
We show in CAFT that class aware frequency transformations of the source sample vs. target sample reduce domain shift in image space, improving model transferability. As a result, current domain adaptation algorithms become easier to adapt under CAFT framework. 
We analyse the effect of class aware frequency transformation in Figure  \ref{fig:acc_curve}(a) and Figure  \ref{fig:acc_curve}(b) using DSLR($D$) $\rightarrow$ Amazon($A$) and Webcam($W$) $\rightarrow$ Amazon($A$) split of Office-31 dataset for DeepCoral \cite{sun2016deep} method. When compared to random transformation, we find that pseudo label based transformation produces better accuracy. For the sake of study, we assume that we have access to the real class label of target samples (Oracle) and compare the outcomes to the random target label and pseudo-label-based approach. The oracle has the highest accuracy (as predicted), followed by the proposed technique and random transformation, both of which are consistent with our hypothesis.
As the quality of pseudo label improves the overall target adaptation performance will improve. In CAFT++, We improve the quality of pseudo labels using ADT2P. We observe the improved performance across the datasets for different domain adaptation methods.

\begin{table*}[!h]
\centering
\caption{Comparison of Precision and Recall for DSAN algorithm} 
\label{tab:precision_recall}
\resizebox{\textwidth}{!}{\begin{tabular}{c|ccccccccccccc}
\toprule
Method & A$\rightarrow$C & A$\rightarrow$P & A$\rightarrow$R & C$\rightarrow$A & C$\rightarrow$P & C$\rightarrow$R & P$\rightarrow$A & P$\rightarrow$C & P$\rightarrow$R & R$\rightarrow$A & R$\rightarrow$C & R$\rightarrow$P & Avg \\
\midrule
Precision CAFT & 67.9 & 78.6 & 85.2 & 64.9 & 73.3 & 74.7 & 64.2 & \textbf{64.4} & 83.4 & 76.2 & 67.1 & 84.4 & 73.7\\
Precision CAFT++ & \textbf{71.2} & \textbf{83.1} & \textbf{89.0} & \textbf{71.1} & \textbf{78.1 }& \textbf{79.5 }& \textbf{68.6 }& 63.6 & \textbf{83.7} & \textbf{77.1} & \textbf{71.5} & \textbf{88.9} & \textbf{77.1}\\
\midrule
Recall CAFT & \textbf{80.8} & \textbf{84.5 }& \textbf{83.7} & \textbf{93.0} & \textbf{95.1 }& \textbf{92.0} & \textbf{94.9} & \textbf{92.1 }& \textbf{94.8} & \textbf{97.5} &\textbf{ 94.4} &\textbf{ 97.9 }& \textbf{91.7}\\
Recall CAFT++ & 80.3 & 78.8 & 79.1 & 81.5 & 85.2 & 82.6 & 87.1 & \textbf{92.1} & 92.1 & 89.8 & 87.2 & 90.9 & 85.6\\

\bottomrule
\end{tabular}}
\end{table*}

\subsection{Effect of ADT2P}

\begin{figure}[!h]
\centering
\includegraphics[width=0.98\linewidth]{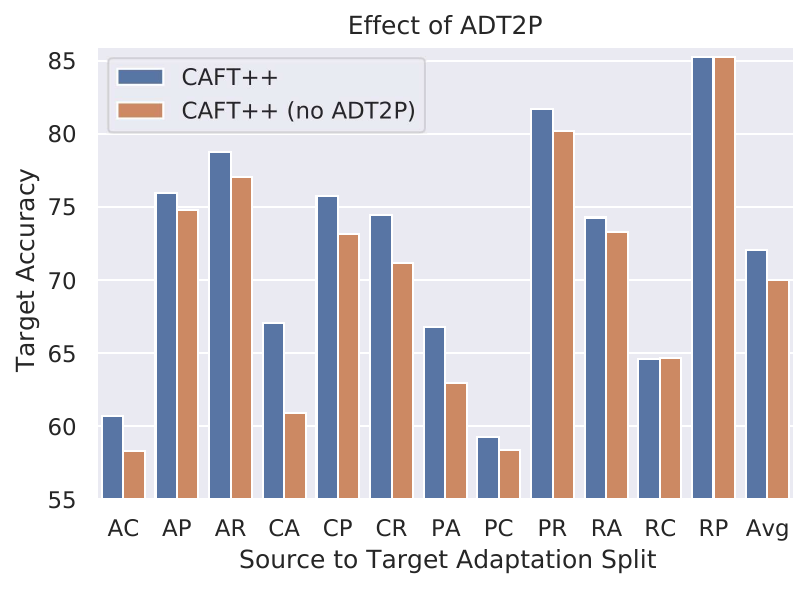}
\caption{\small This plot shows the effect of ADT2P in CAFT++. We can observe that the performance of CAFT++ drops when ADT2P based pseudo label filtering is not used.}
\label{fig:abla_no_adt2p}
\end{figure}

We perform ablation for analysing the effect of proposed pseudo labelling filtering method ADT2P and show the results in Figure  \ref{fig:abla_no_adt2p}. In original CAFT++, the pseudo labels are filtered using ADT2P and we use these clean labels for low-frequency swapping and fine-tuning. In order to analyse ADT2P effect, we remove this is component and perform the overall experiment using threshold based pseudo labelling. We show the result for the same in Figure  \ref{fig:abla_no_adt2p} for each split of Office-Home dataset. CAFT++ with ADT2P performs better for almost all the splits with an average gain across all the splits as \textbf{2.1\%}. 

\noindent \textbf{Compare with Existing Baseline:} In order to show the effectiveness of ADT2P, we compare its performance against similar uncertainty quantification baseline. We replace the ADT2P with Entropy and fit two component Gaussian Mixture Model to separate noisy and clean labels. From Table. \ref{tab:top2_vs_entropy}, we can observe that the performance of CAFT++ drops when we replace ADT2P with Entropy term (decreases by $1\%$ for Office-Home dataset). Therefore, proposed ADT2P can be a better option for quantifying uncertainty in neural network prediction for classification tasks in unsupervised domain adaptation. 

\begin{table*}[!htpb]
\centering
\caption{Comparison between our method vs using entropy in CAFT++} 
\label{tab:top2_vs_entropy}
\resizebox{\textwidth}{!}{\begin{tabular}{c|c|ccccccccccccc}
\toprule
Metric & Method & A$\rightarrow$C & A$\rightarrow$P & A$\rightarrow$R & C$\rightarrow$A & C$\rightarrow$P & C$\rightarrow$R & P$\rightarrow$A & P$\rightarrow$C & P$\rightarrow$R & R$\rightarrow$A & R$\rightarrow$C & R$\rightarrow$P & Avg \\
\midrule
Target & DSAN + CAFT++ (Entropy) & 59.6 & 72.0 & 77.9 & 65.9 & 75.6 & 73.8 & 66.3 & 58.8 & 81.1 & 73.9 & 63.3 & 84.6 & 71.1\\
Accuracy & DSAN + CAFT++ (Ours) & \textbf{60.7} & \textbf{76.0} & \textbf{78.8} &
\textbf{67.1} & \textbf{75.8} & \textbf{74.5} & 
\textbf{66.8} & \textbf{59.3} & \textbf{81.7} & \textbf{74.3} & \textbf{64.6} & \textbf{85.3} & \textbf{72.1}\\

\midrule

Pseudo Label & DSAN + CAFT++ (Entropy) & 69.8 & 78.6 & 87.3 & 69.1 & 77.8 & 79.1 & \textbf{68.6} & 63.1 & 83.2 & 76.3 & 69.2 & 88.0 & 75.8\\
Accuracy & DSAN + CAFT++ (Ours) & \textbf{71.2} & \textbf{83.1} & \textbf{89.0} & \textbf{71.1} & \textbf{78.1} & \textbf{79.5} & \textbf{68.6} & \textbf{63.6} & \textbf{83.7} & \textbf{77.1} & \textbf{71.5} & \textbf{88.9} & \textbf{77.1}\\
\bottomrule
\end{tabular}}
\end{table*}

\begin{table*}[!htpb]
\centering
\caption{Comparison between Top-1 and Top-2 Accuracy} 
\label{tab:top1_top2}
\resizebox{\textwidth}{!}{\begin{tabular}{c|ccccccccccccc}
\toprule
Method & A$\rightarrow$C & A$\rightarrow$P & A$\rightarrow$R & C$\rightarrow$A & C$\rightarrow$P & C$\rightarrow$R & P$\rightarrow$A & P$\rightarrow$C & P$\rightarrow$R & R$\rightarrow$A & R$\rightarrow$C & R$\rightarrow$P & Avg \\
\midrule
MCC + CAFT++ (Top-1) & 59.0 & 80.2 & 83.3 & 67.2 & 78.5 & 79.5 & 67.7 & 54.5 & 82.2 & 74.9 & 61.3 & 85.4 & 72.8\\
MCC + CAFT++ (Top-2) & 69.9 & 88.4 & 91.8 & 78.7 & 87.3 & 87.9 & 78.4 & 64.9 & 90.6 & 82.9 & 72.6 & 92.9 & 82.2\\

\bottomrule
\end{tabular}}
\end{table*}

\subsection{Augmentation's Impact}
\begin{figure}[!h]
\centering
\includegraphics[width=0.98\linewidth]{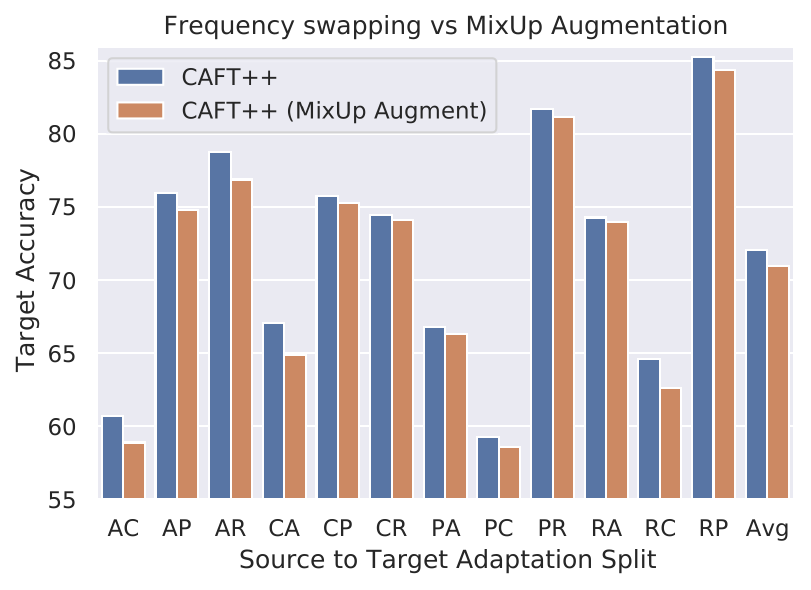}
\caption{\small This plot compares the effect of Fourier Transformation based swapping and MixUp based augmentation. }
\label{fig:abla_mixup}
\end{figure}
\noindent We conduct an ablation for analysing the effect of Fourier transformation based augmentation used in CAFT++ compared to one of the popular MixUp \cite{zhang2018mixup} based augmentation strategy. In MixUp,  we take the convex combination of source and target images. The results in Figure  \ref{fig:abla_mixup} empirically show that our proposed methods perform well in all splits of Office-Home dataset. This analysis confirms the superiority of the Fourier based augmentation in our proposed approach.

\subsection{Effect of Fine-tuning}
\begin{figure}[!h]
\centering
\includegraphics[width=0.98\linewidth]{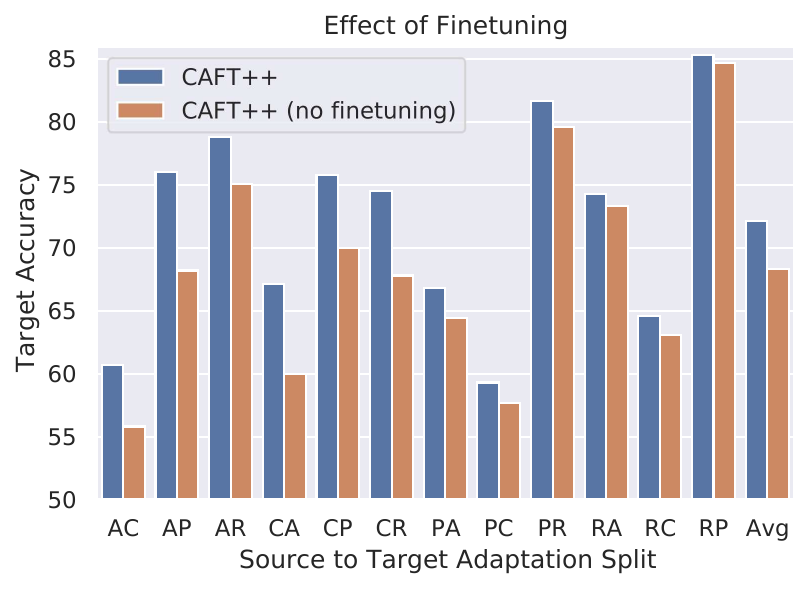}
\caption{\small This plot analyzes the the effect of model parameter fine-tuning using pseudo label based cross entropy loss.}
\label{fig:abla_no_finetune}
\end{figure}
\noindent We observe that the pseudo label filtering using ADT2P gives good percentage of clean labels therefore, we propose to fine-tune the adaptation network using the pseudo-label based cross entropy loss. In this ablation, we analyse the effect of its presence and absence (Figure  \ref{fig:abla_no_finetune}). We can clearly see that fine-tuning using pseudo label based cross entropy loss improves the overall performance.
\subsection{Effect of Original Source}
\begin{figure}[!h]
\centering
\includegraphics[width=0.98\linewidth]{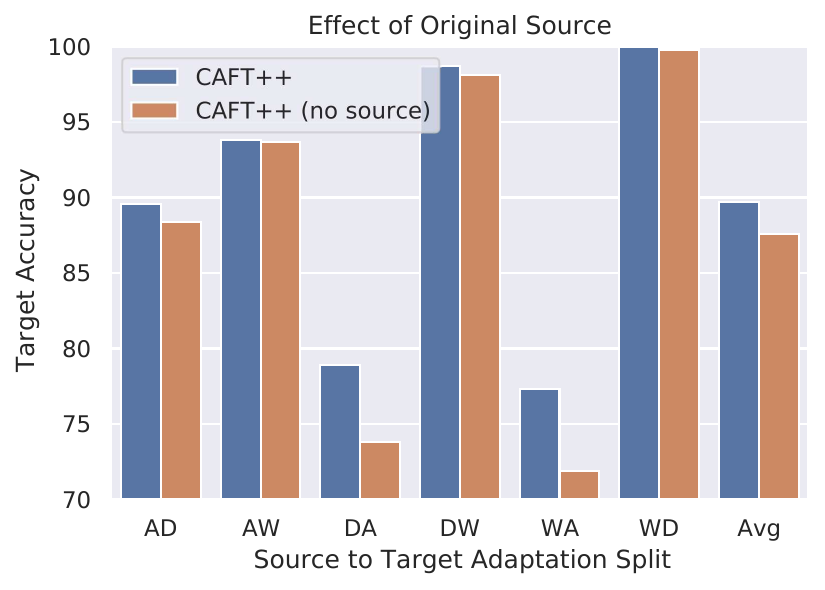}
\caption{\small The effect of using original source along with transformed source during adaptation.}
\label{fig:abla_without_source}
\end{figure}
\noindent We use the transformed source along with the original source in our proposed approach. In this ablation, we analyse the effect of presence and absence of original source sample during adaptation in Fig \ref{fig:abla_without_source}. Model performance drops in the absence of original source. The drop is significant in case of Dslr to Amazon and Webacm to Amazon because transferring style from Amazon to Dslr and Webcam is difficult due to absence of background information in Amazon dataset and it leads to higher degree of artifacts. This experiments also help us to establish that the class-discriminative ability of the transformed source features drops when model transferability improves. We perform this ablation on all the splits of Office-31 dataset.

\subsection{Comparison of different Weight Initialization}

In this ablation study, we analyse the effect of different weight initialization schemes. Previously in CAFT \cite{Kumar_2021_ICCV}, we initialize the experiment with ImageNet trained weights. Due to this reason, during the start of training the pseudo labels will be noisy. We also know the model convergence depends on the weight initialization \cite{li2020comparison}. To have a better initialization, in our current proposal we initialize the training from trained adapted network which results in improved model performance.

\begin{figure}[!htp]
\centering
\centering
\includegraphics[width=\linewidth]{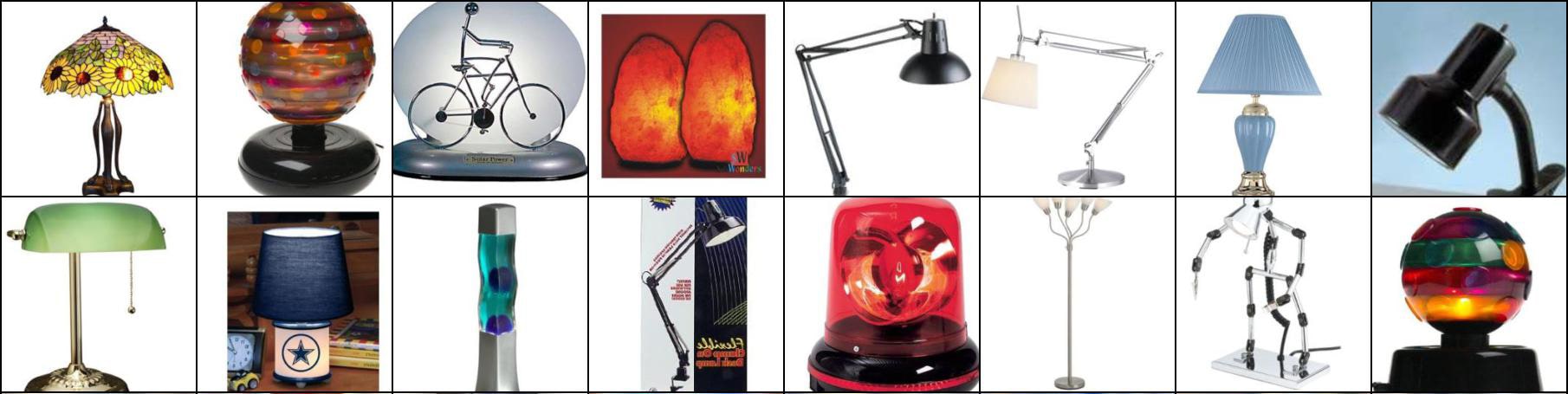}
\caption{ Source (\textbf{\textit{Amazon}}) images }
\label{fig:src}

\end{figure}
\begin{figure}[!htp]
\centering

\centering
\includegraphics[width=\linewidth]{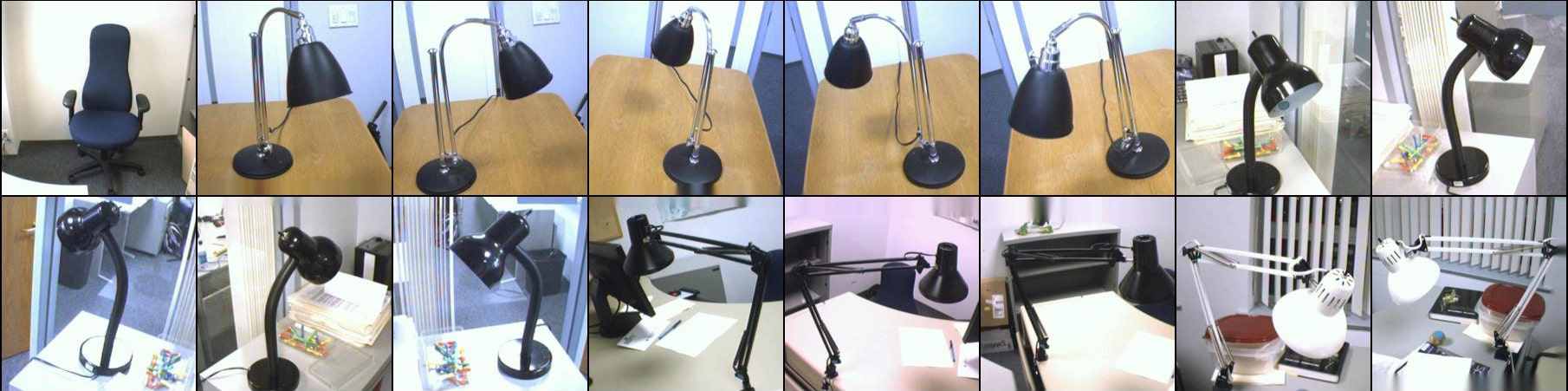}
\caption{ Target (\textbf{\textit{Webcam}}) images }
\label{fig:tgt}

\end{figure}
\begin{figure}[!htp]
\centering

\centering
\includegraphics[width=\linewidth]{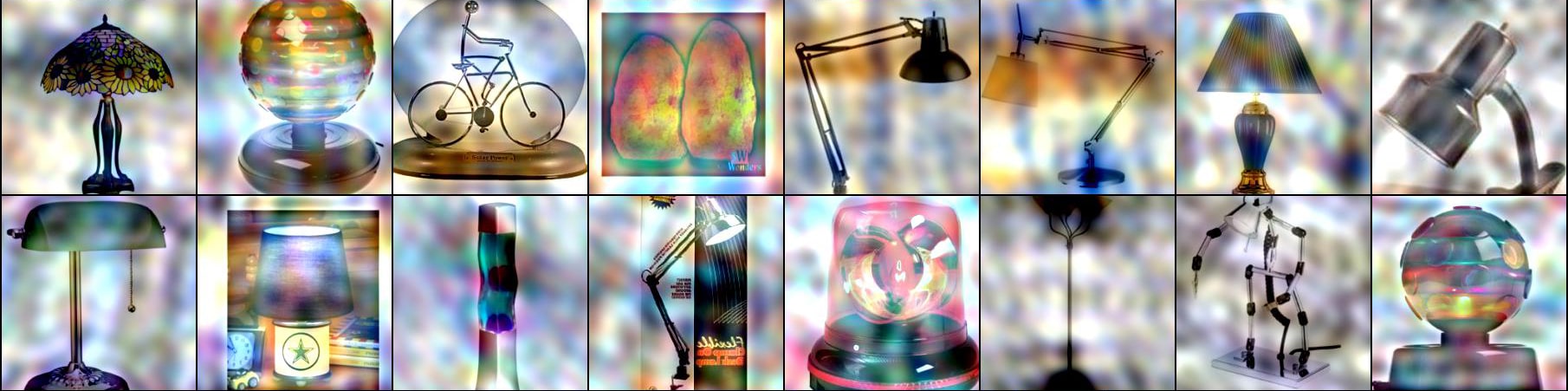}
\caption{ Source (\textbf{\textit{Transformed Amazon}}) images }
\label{fig:s_hat_im}

\end{figure}
\subsection{Analysis of generated Pseudo Label}

\begin{figure}[!h]
\centering
\includegraphics[width=0.98\linewidth]{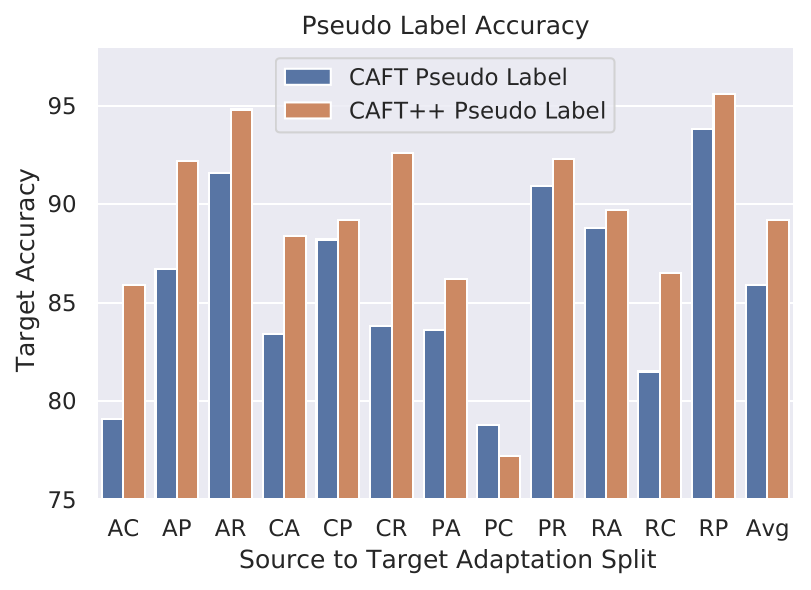}
\caption{\small The above bar plot compares the pseudo label accuracy of CAFT and CAFT++. We can observe that CAFT++ always yields better pseudo label accuracy than compared to CAFT.}
\label{fig:abla_pseudo}
\end{figure}
\begin{figure}[b]
\centering

\centering
\includegraphics[width=\linewidth]{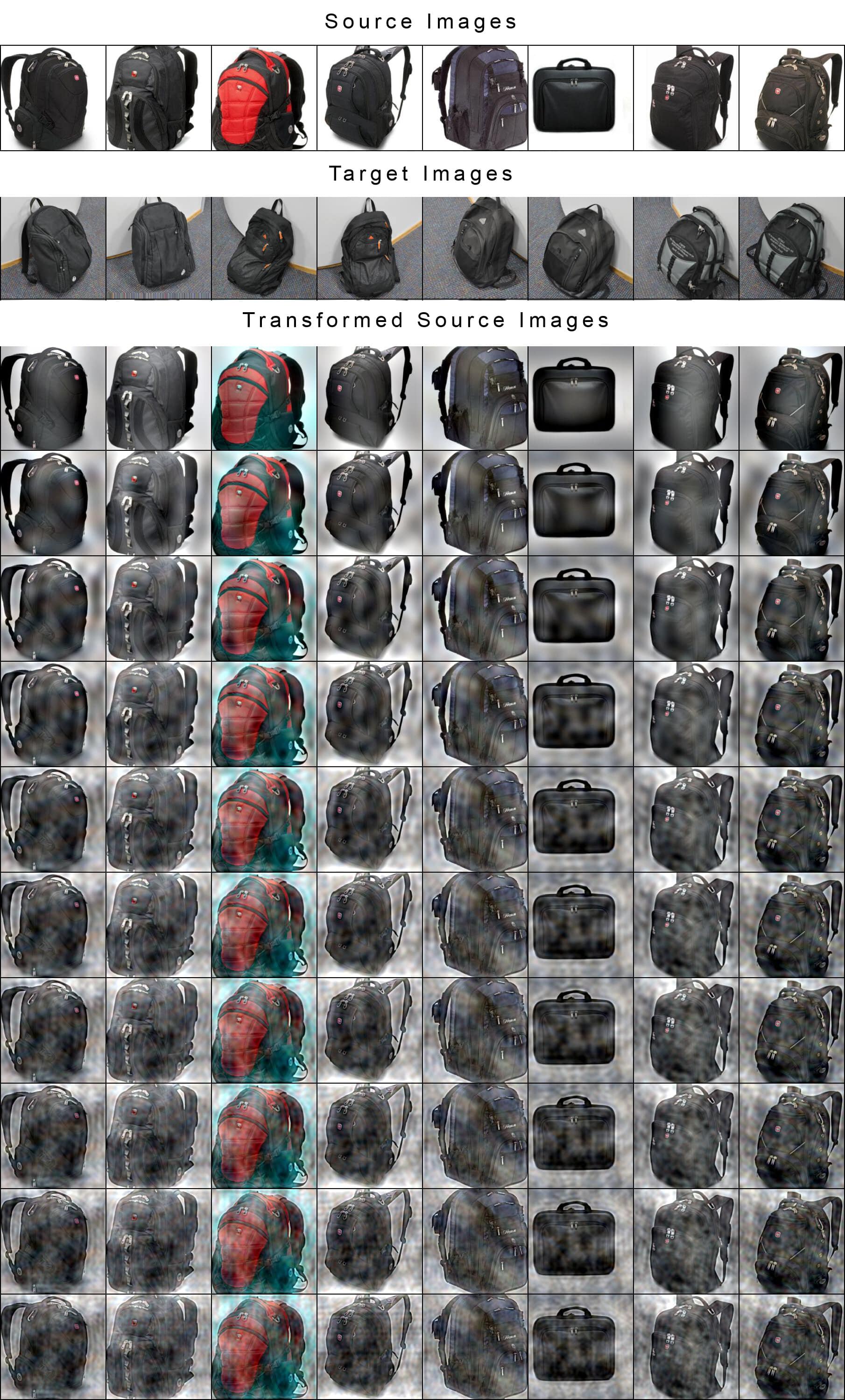}
\caption{ \textbf{Frequency Transformation Effect for different window sizes}. The top row represents the source domain images taken from Amazon dataset. The second row represents the target domain image taken from DSLR dataset. Each row from row 3 onwards represents transformed source image with increasing window size from 1\% up to 10\% with step size of 1\%. As evident, for very low frequency window size, transformation is very weak, while for very high value (near 10\%) undesirable artifacts start to appear in the transformed image. Hence we need to find an intermediate window size for best results. We found that 4\% window size gives the best results for Office-31.}
\label{fig:transformations}

\end{figure}

The accuracy of pseudo label is one of the key components of our proposed approach. We calculate and compare the pseudo label accuracy for CAFT and CAFT++ applied on top of DSAN \cite{zhu2020deep} with Office-Home dataset. We can observe from Figure  \ref{fig:abla_pseudo} that the pseudo label accuracy in case of CAFT++ is improved compared to CAFT. This analysis confirms the effectiveness of our proposed pseudo label algorithm.
\begin{figure*}[!htp]
    \centering
    \begin{minipage}{0.33\textwidth}
        \centering
        \includegraphics[width=1\linewidth]{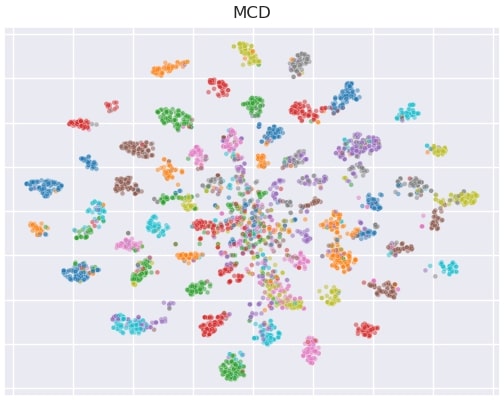}
    \end{minipage}%
    \begin{minipage}{0.33\textwidth}
        \centering
        \includegraphics[width=1\linewidth]{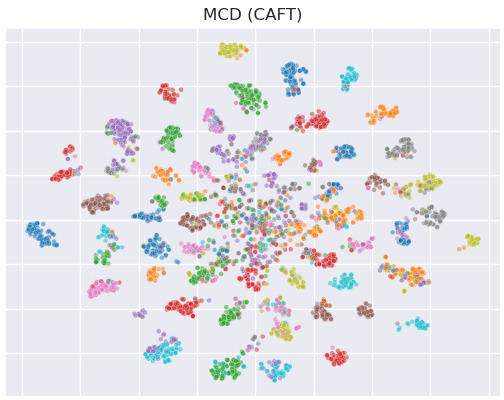}
    \end{minipage}
    \begin{minipage}{0.33\textwidth}
        \centering
        \includegraphics[width=1\linewidth]{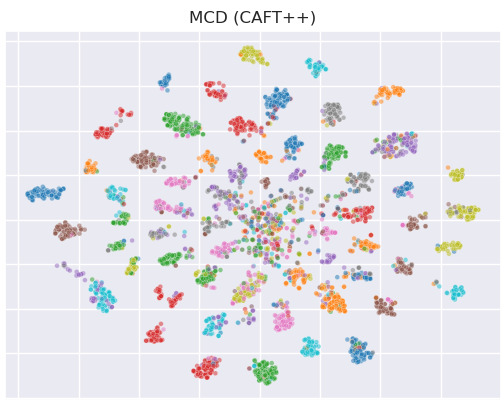}
    \end{minipage}
    \vspace{1mm}
    \caption{\textbf{t-SNE plots}. The above plots compare how well the target is aligned for MCD, MCD + CAFT, MCD +CAFT++ algorithm. Each color represents one of the classes from Office-Home dataset.}
    \label{fig:t-SNE}
\end{figure*}
\subsection{Ablation for selecting window size}

\begin{figure}[!h]
\centering
\includegraphics[width=0.8\linewidth]{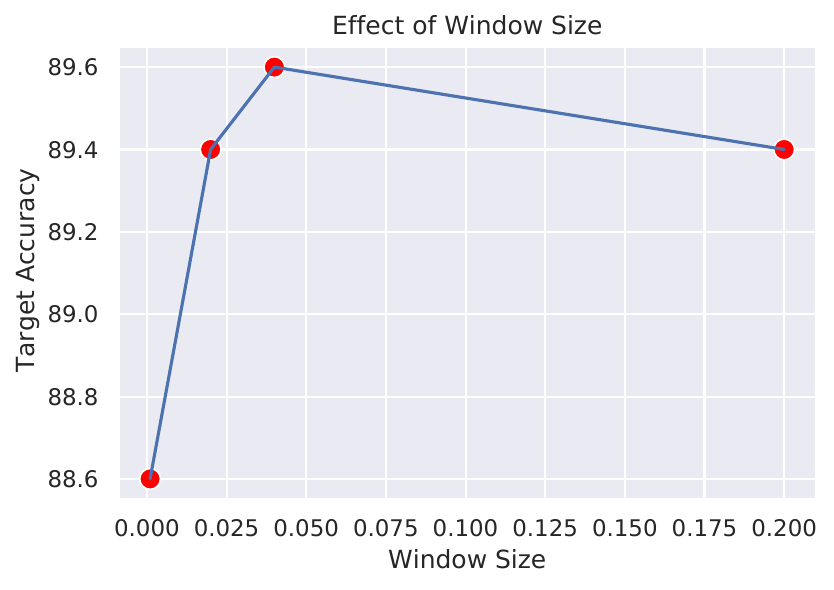}
\caption{\small This ablation shows the effect of selecting various window sizes for frequency replacement.}
\label{fig:abla_window_effect}
\end{figure}
We perform ablation for selecting a hyper-parameter i.e, window size which we use for selecting and swapping the low frequency component. We perform this experiment on Amazon to Dslr of Office-31 dataset. We achieve the best performance when the window size is 0.04 in this case (see Figure \ref{fig:abla_window_effect}). We also observe that the performance remains within the same range for  small variation in the window size, which makes the proposed approach less sensitive to this hyper-parameter.

\section{Qualitative Analysis}

\subsection{Transformed source and target}

We provide qualitative results using class aware Fourier transformation. Figure  \ref{fig:src} shows the source images taken from Amazon domain of Office-31 dataset. Figure  \ref{fig:tgt} shows target images taken from Webcam domain and Figure  \ref{fig:s_hat_im} shows transformed source samples using CAFT++. We can observe that the transformed Amazon source looks closer to the target compared to the original Amazon source.

We also show the transformed source samples by varying the percentage of low frequency components of the source to be transferred. Figure  \ref{fig:transformations} demonstrate the qualitative results for source transformation samples. We observe that if we keep increasing the percentage of low frequency components to be swapped, transformed source sample starts loosing the class-discriminative features, which will reduce the classification performance.

\subsection{Visualization using t-SNE plots}

Using t-SNE, We analyse the discriminative feature alignment for source and target dataset features taken from the output of adapted model feature extractor. We experiment with RealWorld $\to$ Product source-target pair of  Office-Home dataset on MCD adaptation method with and without CAFT++. We can observe the alignment of discriminative features in Figure \ref{fig:t-SNE}, which looks more separable in projected space for MCD + CAFT++. A better clustering of features in the latent space implies well separation of class-discriminative features. It should result in better model performance. We have also added different colors for different classes. It helps us to visualize the goodness of clusters in two domains and shows the alignment of corresponding classes in two domains.

\section{Conclusion}
Using classic image processing methods such as Fourier transformation, this work looked at limiting negative transfer and achieving explicit domain shift reduction in image space for improved transferability. Our novel pseudo label based filtering scheme was successfully able to  filter out clean and noisy pseudo labels. In the proposed CAFT++ framework, with the use of clean pseudo labels, we investigated the efficacy and improvement over the Class Aware Fourier Transform (CAFT) from source to target data samples. We empirically evaluated our proposed hypothesis by performing many ablations. In comparison to existing domain adaptation baseline approaches, the suggested strategy produces better or equivalent outcomes. Proposed framework CAFT++ acts as an aid to domain adaptation and is independent of the UDA algorithm. As a result, it may easily be included in any existing domain adaptation methods to increase their current performance. Unlike adversarial network-based style transfer approaches for generating intermediate domain images, our proposed approach does not need adversarial network-based style transfer therefore, our proposed approach is computationally inexpensive.

\noindent \textbf{Acknowledgement:} This work is partially supported by a Young Scientist Research Award (Sanction no. 59/20/11/2020-BRNS) to Anirban Chakraborty from DAE-BRNS, India.

\bibliography{cpp-bibliography}
\end{document}